\documentclass{article}

\usepackage{microtype}
\usepackage{graphicx}
\usepackage{subcaption}
\usepackage{booktabs} % for professional tables

\usepackage{hyperref}
\usepackage{bbm}

\usepackage[preprint]{icml2026}

\usepackage{enumitem}
\usepackage{amsmath}
\usepackage{amssymb}
\usepackage{mathtools}
\usepackage{amsthm}

\usepackage[capitalize,noabbrev]{cleveref}

\theoremstyle{plain}

\theoremstyle{definition}

\theoremstyle{remark}

\usepackage[textsize=tiny]{todonotes}

\icmltitlerunning{Picking the Right Specialist}

\begin{document}

\twocolumn[
  \icmltitle{Picking the Right Specialist: Attentive Neural Process-based Selection of Task-Specialized Models as Tools for Agentic Healthcare Systems}

  \begin{icmlauthorlist}
\icmlauthor{Pramit Saha}{yyy}
\icmlauthor{Joshua Strong}{yyy}
\icmlauthor{Mohammad Alsharid}{xxx}
\icmlauthor{Divyanshu Mishra}{yyy}
\icmlauthor{J. Alison Noble}{yyy}

\end{icmlauthorlist}

\icmlaffiliation{yyy}{Department of Engineering Science, University of Oxford, United Kingdom}
\icmlaffiliation{xxx}{Department of Computer Science, Khalifa University, Abu Dhabi, United Arab Emirates}

\icmlcorrespondingauthor{Pramit Saha}{pramit.saha@eng.ox.ac.uk}

  \icmlkeywords{Machine Learning, ICML}

%   \printAffiliationsAndNotice{}

  \vskip 0.3in
]

\printAffiliationsAndNotice{}  % leave blank if no need to mention equal contribution

\begin{abstract}
Task-specialized models form the backbone of agentic healthcare systems, enabling the agents to answer clinical queries across tasks such as disease diagnosis, localization, and report generation. Yet, for a given task, a single “best” model rarely exists. In practice, each task is better served by multiple competing specialist models where different models excel on different data samples. 
As a result, for any given query, agents must reliably select the right specialist model from a heterogeneous pool of tool candidates. To this end, we introduce \textbf{ToolSelect}, which adaptively learns model selection for tools by minimizing a population risk over sampled specialist tool candidates using a consistent surrogate of the task-conditional selection loss. Concretely, we propose an Attentive Neural Process–based selector conditioned on the query and per-model behavioral summaries to choose among the specialist models. Motivated by the absence of any established testbed, we, for the first time, introduce an agentic Chest X-ray environment equipped with a diverse suite of task-specialized models (17 disease detection, 19 report generation, 6 visual grounding, and 13 VQA) and develop \textbf{ToolSelectBench}, a benchmark of 1448 queries. Our results demonstrate that ToolSelect consistently outperforms 10 SOTA methods across four different task families.

\end{abstract}

\section{Introduction}
\begin{figure}
    \centering
\includegraphics[width=0.97\columnwidth]{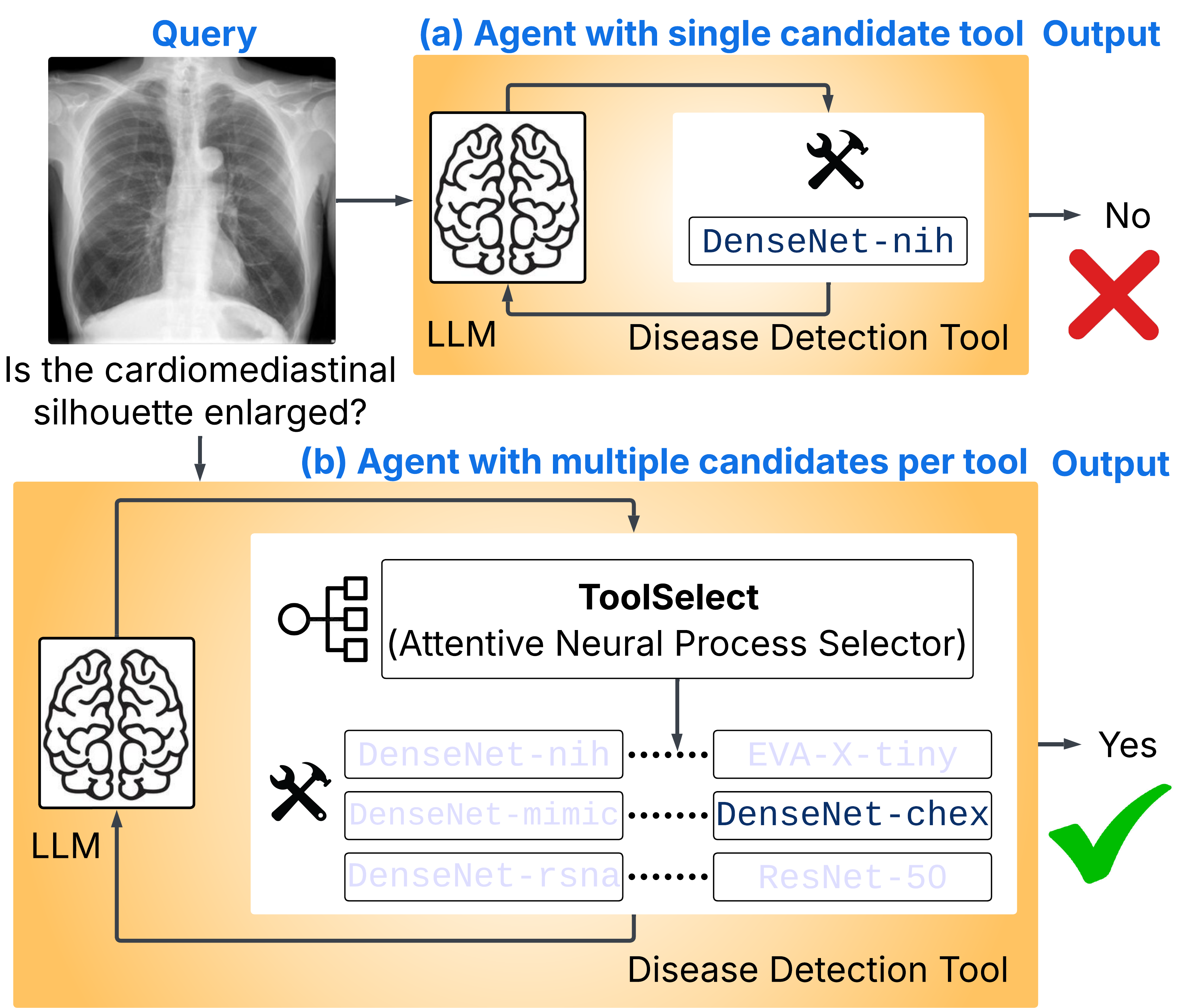}
    \caption{\textit{Why model selection for tools matters in agentic systems?} \textbf{Top (a):} A naive setup with a single fixed specialist model per tool fails under domain and label-space mismatch. \textbf{Bottom (b):} Our approach equips tools with multiple specialist candidates and uses \textbf{ToolSelect to adaptively select appropriate model for a given query}, yielding more accurate clinical answers than (a).}
\label{fig:11}
\end{figure}
\begin{figure*}[t]
    \centering
\includegraphics[width=2.1\columnwidth]{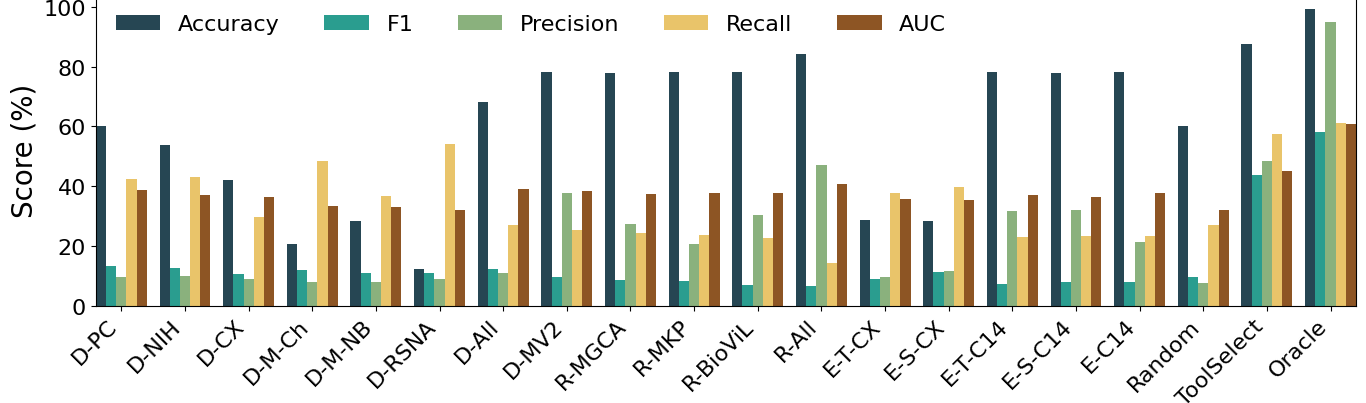}
\caption{Performance comparison of \textbf{ToolSelect} against 18 task-specialized chest X-ray disease-detection models. Individual specialists exhibit low performance and substantial performance variability. The \textbf{Oracle} upper bound indicates that selecting the appropriate specialist per case can yield large gains, whereas \textbf{Random} selection often degrades performance, sometimes below that of individual models. \textbf{ToolSelect} consistently improves over individual specialists and random selection, aiming to close the gap to the Oracle.}
\label{fig:11}
\end{figure*}
Recent advances in large language models (LLMs) \cite{thirunavukarasu2023large,naveed2025comprehensive} have enabled agentic systems \cite{naveed2025comprehensive, wang2024survey, guo2024large} that can plan, reason, and interact with external tools, raising the prospect of more capable AI assistants in healthcare \cite{qiu2024llm, wang2025survey}. However, in clinical settings, LLMs alone remain insufficient \cite{mofrad2025limitations}: they often lack up-to-date medical knowledge and expertise, struggle with domain-specific terminology and reasoning, and cannot directly solve complex clinical tasks based on high-dimensional data such as medical images and videos. As a result, agentic healthcare systems increasingly rely on task-specialized models as external tools to answer clinical queries precisely \cite{fallahpour2025medrax, tang2025endoagent, huang2026surgical}. These specialist models encapsulate domain expertise that general-purpose LLMs cannot reliably acquire through language-only training, making tool use a foundational component of practical clinical agents.

In real-world healthcare, a single tool per task is rarely sufficient. Clinical deployment faces substantial variability across hospitals, scanners, acquisition protocols, and patient populations, leading to pronounced domain shifts that can substantially degrade model reliability \cite{maleki2022generalizability, sanner2021reliable,musa2025systematic}. At the same time, training datasets differ in label definitions, granularity, and prevalence, creating label-space and label-distribution gaps that make “one-size-fits-all” training unrealistic. These issues are compounded by variable annotation quality, partial label coverage, heterogeneous reporting styles, and dataset-specific biases, as a result of which, models that perform well in one setting often fail to transfer to another. Therefore, practical agentic healthcare systems must maintain multiple task-specialized models as tools, covering different data regimes, label spaces, and clinical subpopulations. This makes tool candidate selection a pressing bottleneck in agentic healthcare systems. In other words, effective clinical performance depends not only on having access to a number of specialist candidate models as tools, but on reliably selecting the right one for each query.

In this paper, we study the problem of query-based tool candidate selection in agentic healthcare systems, \textit{i.e.},  \textit{given a clinical query and a pool of specialized black-box models available for the selected task, we ask: How can an agent reliably choose the most appropriate specialist for that specific input?} To address this, we introduce \textbf{ToolSelect}, a learning-based selection framework that performs query-based adaptive model selection within a task by minimizing a population risk over sampled candidate panels using a consistent surrogate of the task-conditional selection loss. ToolSelect represents each candidate model through compact behavioral summaries and employs an Attentive Neural Process selector conditioned on the query, enabling principled selection among heterogeneous specialists while keeping all underlying tools frozen (see Fig. 3).

To support rigorous evaluation in a realistic medical setting, we introduce an agentic chest X-ray environment with an LLM-based agent core and a diverse suite of task-specialized models spanning disease detection, report generation, visual grounding, and visual question answering (VQA). Subsequently, we develop a benchmark for investigating tool candidate selection called \textbf{ToolSelectBench} that instantiates the practical challenges motivating this work including domain shifts, mismatched label spaces, and partial label support within a single end-to-end agent workflow This allows a systematic comparison of routing strategies under controlled but clinically meaningful tasks. Across ToolSelectBench, our proposed method consistently improves agent performance over a wide range of baselines.
Our primary contributions can be summed up as follows: 
\begin{itemize}[leftmargin=0cm, itemsep=2pt, topsep=2pt]
\item \textbf{Task, Framework, and Dataset contribution:} To the best of our knowledge, this is the first work to highlight (a) the importance of maintaining {multiple} specialist models as tool candidates in agentic healthcare systems and (b) the need for a {query-conditioned} tool-selection module that routes each request to an appropriate specialist. To enable reproducible study of this problem, we develop a testbed of 1448 Chest X-ray and question-answer triplets as well as a novel chest X-ray-based agentic environment comprising an inference-ready tool candidate pool spanning four task families: 17 disease-detection models, 19 report-generation models, 6 visual-grounding models, and 13 VQA models. These tools cover a broad range of diseases and clinical tasks and were either trained independently (from scratch or fine-tuned) or collected from diverse public sources. More details in \S 3.
\item \textbf{Technical contribution:} We first formalize {multi-task query-based tool candidate selection} in agentic healthcare as a {population-risk minimization} problem over candidate models, each with {partial task support}. Building on this, we propose \textbf{ToolSelect}, a {query-conditioned} model selection method based on {Attentive Neural Processes (ANP)} and {per-model behavioral summaries} to select among available {frozen specialist models}. While we do not introduce new theorems, this work shows the importance of ANP-based specialist model selection and comp-sum loss along with corresponding theoretical justification (See Suppl. B, C).

\item \textbf{Empirical contribution:} We first investigate the standalone performance of 55 specialist models in our tool pool and release their outputs and responses to facilitate reuse and consistent evaluation without re-running training or inference. Additionally, we benchmark ToolSelect for each of the four Chest X-Ray-based task families against \textbf{12} diverse baselines spanning heuristic, ML-based selectors as well as LLM routing strategies adapted to tool selection setting. 
     
\end{itemize}

\begin{figure*}
    \centering
\includegraphics[width=1.36\columnwidth]{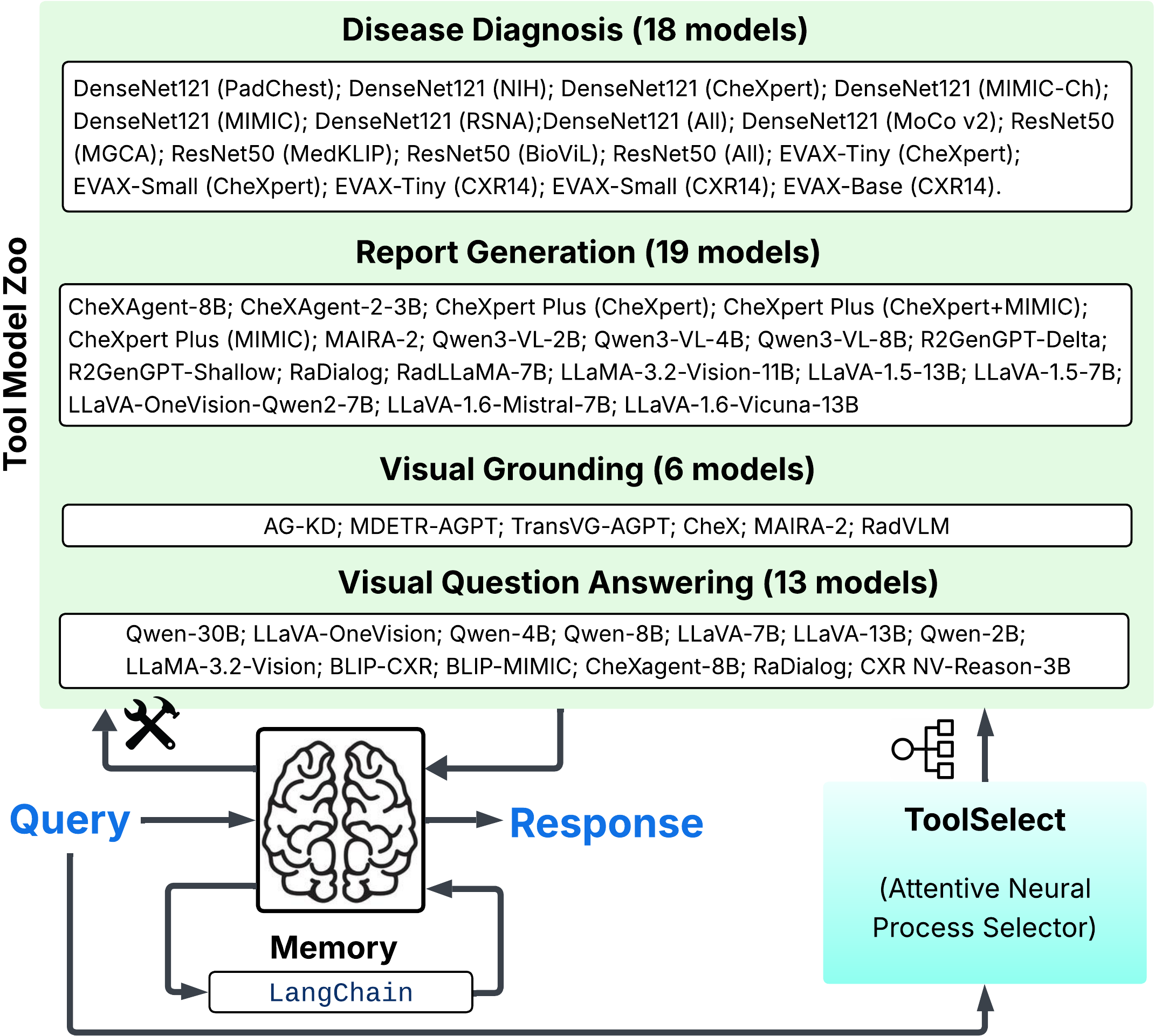}
    \caption{Architecture of our agentic framework. The system follows a ReAct-style loop, combining short-term memory (LangChain) with a heterogeneous tool model zoo incorporating fundamental Chest X-Ray-based tasks required to process user queries. Our proposed ToolSelect module is integrated to perform query-conditioned tool candidate (i.e., specialist model) selection.}
\label{fig:11}
\end{figure*}

\begin{figure*}[t]
    \centering
\includegraphics[width=2\columnwidth]{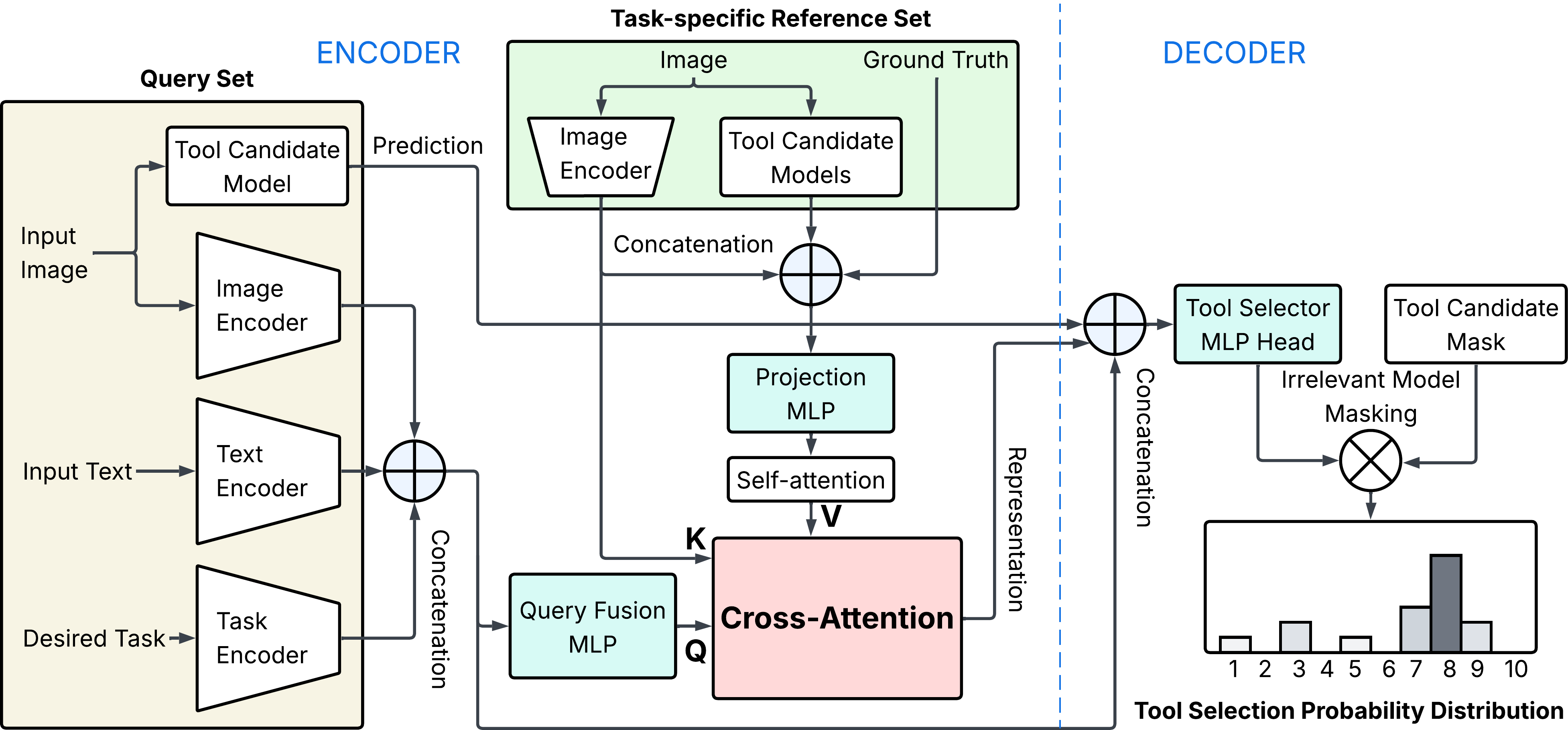}
    \caption{Overview of the proposed \textbf{ToolSelect} architecture. Given a multimodal query (image, text, and task), query features are fused and used to attend over task-specific per-tool reference sets that summarize each candidate tool’s empirical behavior. Cross-attention produces a query-conditioned representation for each tool, which is combined with the tool’s prediction and passed to a selector network. Tools that do not support the queried task are masked out before producing the final tool selection probability distribution.}
\label{fig:11}
\end{figure*}

\section{Methodology}
\subsection{Problem formulation}
\textbf{Input and tasks.}
Each user query is a multimodal prompt $p=(x,q)$ consisting of an image $x\in\mathcal{X}$ and a natural-language instruction $q$ (e.g., ``Is there evidence of pneumonia in this chest X-ray?''). A task selector (LLM orchestrator) deterministically maps the prompt to a task $t=\tau(p)\in\mathcal{T}$ (e.g., \texttt{disease detection}, \texttt{visual grounding}, \texttt{report generation}, \textit{etc.}).

\noindent\textbf{Population of tools per task.}
For each task $t$, we assume a population distribution $P_t(E)$ over tool candidates (``specialists''). At inference time, the system observes an available panel: $S_t=(E_1,\ldots,E_m)\ \sim\ P_t(E)^m $ with $m\ge2$ (i.i.d.\ draws for simplicity). Each tool $E_i$ provides a task-specific prediction rule $g_E^t$ (e.g., class logits for disease detection) evaluated on inputs $x$. Tools may exhibit {partial label-conditional support} within a task, \textit{i.e.},  for output components they cannot reliably predict, they emit a distinguished null prediction, implemented as a constant score in our experiments. Such tools still appear in $S_t$, but are automatically {masked out} for unsupported components during selection and loss computation to avoid interference.

\noindent\textbf{Ground-truth outputs and costs.}
Let $y^t\in\mathcal{Y}_t$ denote the ground-truth output for task $t$ (e.g., a disease label, correct MCQ option, an original report, or an expert-annotated bounding box). We evaluate each tool $E_i$ using a bounded, task-specific cost $c_E^t(x,y^t)\in[0,1]$ instantiated as follows:
\begin{itemize}[leftmargin=0cm, itemsep=2pt, topsep=2pt]
\item \textbf{Disease Detection:} $c_E^t(x,y)=\mathbbm{1}\!\left[\arg\max g_E^t(x)\neq y\right]$, or calibrated cross-entropy clipped to $[0,1]$.
\item \textbf{Visual Grounding:} 
$c_E^t(x,y)=1-\mathrm{IoU}\!\left(\hat{y}_E(x),y\right)$, where intersection over union ($\mathrm{IoU}$) is evaluated between predicted and ground-truth regions for each instance.
\item \textbf{Report Generation:} 
$c_E^t(x,y)=1-\mathrm{F1}_{\text{RadGraph}}\!\left(\hat{y}_E(x),y\right)$, 
which measures structured clinical correctness between the generated report and ground truth per instance.
\item \textbf{VQA:}
$c_E^t(x,y)=\mathbbm{1}\!\left[\hat{y}_E(x,q)\neq y\right]$ 
where $\hat{y}_E(x,q)$ is the option selected by VLM for the multiple-choice question $q$.

\end{itemize}
\noindent\textbf{Partial support and abstention.}
Each tool may not provide meaningful predictions for all output components under a given task. We represent such abstentions via a distinguished null prediction and define a task-conditional validity indicator $\mathbbm{1}\left[(x,t)\in\sigma_E\right]$, where $\sigma_E$ denotes the support of tool $E$. The cost $c_E^t(x,y^t)$ is evaluated only on supported components. Tools outside the effective support are masked out for corresponding selection decision and contribute neither to loss nor to normalization of selection probabilities.

\subsection{Tool candidate model selection }
\textbf{Selector.}
Given a prompt–task–panel tuple $(p,t,S_t)$, we define a parametric selector $r_\theta$ that assigns a real-valued score to each tool candidate in the available panel $S_t$. Formally,
\begin{align*}
&r_\theta : (\mathcal{X}\times\mathcal{Q}) \times \mathcal{E}^m \rightarrow \mathbb{R}^m \\
&(p,S_t) \mapsto \big(r_\theta(p,1;S_t),\ldots,r_\theta(p,m;S_t)\big)
\end{align*}
where $r_\theta(p,j;S_t)$ denotes the score assigned to the $j$-th tool $E_j$ in the panel, conditioned on $p$ and full panel context $S_t$. We convert these scores into selection probabilities $\pi_\theta(j \mid p,S_t)$ via a masked softmax:
\begin{equation*}
\pi_\theta(j \mid p,S_t)
= \frac{\exp\!\left(r_\theta(p,j;S_t)\right)\,\mathbbm{1}\left[(x,t)\in\sigma_{E_j}\right]}
{\sum_{k=1}^m \exp\!\left(r_\theta(p,k;S_t)\right)\,\mathbbm{1}\left[(x,t)\in\sigma_{E_k}\right]}
\end{equation*}
where $\mathbbm{1}[(x,t)\in\sigma_{E_j}]$ is a task-conditional validity indicator that masks out tools that do not support the current input–task pair. At inference time, the selected tool is
\[
\hat{j}(p,S_t)=\arg\max_{j}\,\pi_\theta(j\mid p,S_t)
\]
\noindent\textbf{Task-conditional selection risk.}
For a single decision with panel $S_t$, the population selection loss is defined as:
\[
L^{t}_{\mathrm{sel}}(r_\theta;p,y^t,S_t)
=\sum_{j=1}^m c_{E_j}^t(x,y^t)\ \mathbbm{1}\!\left[\hat{j}(p,S_t)=j\right]
\]
The population risk averages over data and panels:
\[
E_{(X,Y^t)\sim D_t,\ S_t\sim P_t(E)^m}\left[L^{t}_{\mathrm{sel}}(r_\theta; (X,q),Y^t,S_t)\right].
\]
\textbf{Comp-sum surrogate loss.}
Directly minimizing $L_{\mathrm{sel}}$ is intractable, as the induced selection rule is combinatorial and non-differentiable. 
We therefore adapt the comp-sum surrogate of the loss following \cite{mao2023cross,mao2023two} to our tool candidate selection setting. 
For any decreasing function $\Psi:[0,1]\!\to\!\mathbb{R}_+$ with $\Psi(1)=0$, we define:
\[
\begin{aligned}
L^{t}_{\Psi}(r_\theta; p,y^t,S_t)
&= \sum_{j:\,E_j \text{ valid for } t} 
\underbrace{\Big(\sum_{j'\neq j} c_{E_{j'}}^t(x,y^t)-m+2\Big)}_{w_j(p,y^t,S_t)} \\
&\quad \times \Psi\!\left(\pi_\theta(j\mid p,S_t)\right)
\end{aligned}
\]

\noindent Consequently, we train the selector by minimizing the task-marginalized objective (with $\lambda_t$ as the task weights):
\[
\mathcal{E}_{\Psi}(r_\theta)
=\sum_{t\in\mathcal{T}} \lambda_t\ 
E_{(X,Y^t)\sim D_t,\ S_t}\!\left[
L^{t}_{\Psi}(r_\theta;(X,q),Y^t,S_t)
\right]
\]
where $\lambda_t$ are the task weights. 
In all experiments, we use the logistic comp-sum loss, $\Psi(u)=-\log u$. 
For $m=2$, the surrogate reduces to a cost-sensitive binary margin form:
$c_{E_2}^t\,\Phi(\Delta)-c_{E_1}^t\,\Phi(-\Delta)$ with $\Delta=r_\theta(p,1)-r_\theta(p,2)$.

\subsection{Attentive Neural Process for tool candidate selection}
\noindent\textbf{Relation to meta-learning with reference sets.}
Our selector follows a {model-based meta-learning} paradigm for learning from tool-specific reference sets. In contrast to
meta-optimization approaches which adapt parameters via gradient updates at test time, we
condition the selector directly on a tool's behavioural reference set through a single forward pass. We instantiate this
conditioning using an Attentive Neural Process (ANP) \cite{kim2019attentive}, enabling query-dependent adaptation without test-time optimization.

\textbf{Label-space alignment.}
Our tool candidates have heterogeneous output spaces $\mathcal{Y}_E^t$ (e.g., different classes, phrase vocabularies or mask resolutions). We maintain a known mapping $\rho_E^t:\ \mathcal{Y}_E^t\to 2^{\mathcal{Y}_t}$ to a canonical task space $\mathcal{Y}_t$. 
For classification, we compute aligned probabilities $\tilde{m}_E^t(x)\in\Delta^{|\mathcal{Y}_t|-1}$ by summing over pre-image labels. When a tool does not support a particular label or component, $\tilde{m}_E^t(x)$ assigns a uniform (uninformative) score, which we treat as a null prediction and mask out during routing and loss computation.

\noindent\textbf{Task-specific reference set (few-shot behavioural descriptors).}
To support query-guided selection, each specialist model $E$ provides a small reference set:
\[
D_E^t=\{d_{E,b}^t\}_{b=1}^{B_t},\qquad 
d_{E,b}^t=(x_b,\,y_b^t,\,m_E^t(x_b)),
\]
which summarizes its behaviour on task $t$. Here $m_E^t(x_b)$ denotes the tool’s aligned prediction and $B_t$ is kept intentionally small (tens of examples; $B_t\in[16,64]$ in our experiments). In practice, $B_t$ should be as large as possible while remaining feasible for tool maintenance (\textit{i.e.}, without incurring undue cost to produce or refresh predictions).

\noindent\textbf{Query and tool encoders.}
We encode the input image, text, and instruction with task-shared backbones ($\phi_x, \phi_q$):
\begin{align*}
&\phi_x:\mathcal{X}\to\mathbb{R}^{d_x},\qquad \phi_q:\mathcal{Q}\to\mathbb{R}^{d_q},\\&
u(p)=U\,[\phi_x(x)\ \|\ \phi_q(q)\,]\in\mathbb{R}^{d_u}.
\end{align*}

\noindent\textbf{Reference encoding.} Each reference element is embedded:

\noindent\makebox[\linewidth][l]{$
\begin{aligned}
t_b
&=W_c\big[\phi_x(x_b)\ \|\ e^t(y_b^t)\ \|\ \rho_m(\tilde{m}_E^t(x_b))\big]\in\mathbb{R}^d, T=[t_1,.,t_{B_t}]
\end{aligned}$}

\noindent\textbf{Reference self-attention and query–reference cross-attention.} To obtain richer representations of the reference set, we next apply self-attention. Given query representation $u(p)$, we apply cross-attention between the query and the reference set:

\noindent\makebox[\linewidth][l]{$
\small
\begin{aligned}
\tilde{T}&=\mathrm{SelfAttn}(T), K=W_K\big[\phi_x(x_1),.,\phi_x(x_{B_t})\big]^\top,V=W_V\,\tilde{T},\\
z_E^t(p)
&=\mathrm{softmax}\!\Big(\frac{(W_Q u(p))K^\top}{\sqrt{d_k}}\Big)\,V
\in \mathbb{R}^{d_v}.
\end{aligned}
$}

The resulting vector $z_E^t(p)$serves as a \textbf{query-dependent tool descriptor}  $\psi_E^t(p)$, emphasizing those reference examples most relevant to the current input. This allows the selector to prefer tools that perform well on inputs similar to the query, even if they are not globally optimal on average.

\noindent\textbf{Selector head.}
We score tool candidate model $E$ as:
\[
r_\theta(p,E;S_t)=g_\theta\!\big(\,[u(p)\ \|\ \psi_E^t(p)\ \|\ \tilde{m}_E^t(x)\ \|\ \eta_E]\,\big),
\]
with a lightweight MLP $g_\theta$ shared across tasks and $\eta_E$ encodes optional tool metadata. The scores are normalized into $\pi_\theta(j\mid p,S_t)$ as described in \S 2.2.

\subsection{Training objective and algorithm}

\noindent\textbf{Overall objective.}
We minimize $\mathcal{E}_{\Psi}(r_\theta)$ (see \S~2.2) together with the regularizers described below. Gradients propagate through the router and the ANP encoders, while tool candidate models $g_E^t$ are kept frozen. See Suppl. B \& C. 
\\
\noindent\textbf{Panel sampling.}
At each SGD step, we sample a task $t\sim\mathrm{Cat}(\lambda)$, a minibatch $\{(x_i,y_i^t,q_i)\}_{i=1}^B\sim D_t$, and, for each $i$, an available tool panel $S_t^{(i)}\sim P_t(E)^m$. We compute the scores $\{r_\theta(p_i,E;S_t^{(i)})\}$, form the selection probabilities $\pi_\theta$, construct weights $w_j$ from costs $c_{E_j}^t(x_i,y_i^t)$, and accumulate the comp-sum loss. 
\\
\noindent\textbf{Regularization.}We employ two regularizers: (i) a \emph{panel entropy} term $\lambda_H\,E[-\sum_j \pi_\theta(j)\log \pi_\theta(j)]$ to discourage overconfident selections early in training, and (ii) a per-tool \emph{coverage head} $s_E^t(p)=\sigma(h([u(p)\|\psi_E^t(p)]))$, trained with targets $(1-c_E^t)$ to approximate the tool’s task-conditional success probability.
\section{Dataset and Experimental Setup}
\subsection{Tasks and Datasets}
We consider four tool sets corresponding to fundamental tasks in chest X-ray--based clinical workflows: (i) disease diagnosis, (ii) report generation, (iii) phrase-level visual grounding, and (iv) multiple-choice VQA for chest X-Ray interpretation.
For disease diagnosis, we construct evaluation queries using two external datasets, Open-I \cite{jaeger2014two} and VinDr-CXR \cite{nguyen2022vindr}, which introduce clinically realistic domain shifts (e.g., scanner, acquisition protocol, and patient population) and label-distribution mismatch relative to the training sources of the candidate models. Similarly, for report generation, we use Open-I and for phrase-level grounding, we use VinDr-CXR. For multiple-choice VQA, we sample 1\% of the ReX-VQA dataset \cite{pal2025rexvqa}. Across all tasks, we follow the standard dataset splits: the training split is used for training the selector, and performance is reported on the corresponding test split. We construct \textbf{ToolSelectBench}, a unified benchmark comprising \textbf{550} disease diagnosis, \textbf{156} report generation, \textbf{394} phrase-level grounding, and \textbf{348} multiple-choice VQA queries, all drawn from the corresponding test set. All benchmark samples, together with the individual predictions of \textbf{55} candidate models, are attached as \textbf{supplementary material} for transparency and reproducibility.
\subsection{Tool Candidate Model Zoo}
\noindent\textbf{Disease Diagnosis:} We collected and/or trained 17 chest X-ray (CXR) models spanning three architectures: (a) DenseNet121, (b) ResNet50, and (c) ViT. The DenseNet121 models \cite{cohen2022torchxrayvision} and their training data are: (i) DenseNet121–PadChest (D-PC), (ii) DenseNet121–NIH-CXR8 (D-NIH), (iii) DenseNet121–CheXpert (D-CX), (iv) DenseNet121–MIMIC-Ch (D-M-Ch), (v) DenseNet121–MIMIC-NB (D-M-NB), (vi) DenseNet121–RSNA-CXR (D-RSNA), (vii) DenseNet121–All-the-above-datasets (D-All), and (viii) DenseNet121–CXR-14, trained via MoCo-v2 (D-MV2). The ResNet50 models are: (i) ResNet50-CXR-14 trained via MGCA (R-MGCA), (ii) ResNet50-CXR14 trained via MedKLIP (RN50-MKP), (iii) ResNet50-CXR-14 trained via BioViL (R-BioViL), and (iv) ResNet50–All (R-All). Finally, our ViT-based models use the EVA-X foundation model \cite{yao2025eva} fine-tuned on respective datasets: (i) EVAX-Tiny–CheXpert (E-T-CX), (ii) EVAX-Small–CheXpert (E-S-CX), (iii) EVAX-Tiny–NIH-CXR14 (E-T-C14), (iv) EVAX-Small–NIH-CXR14 (E-S-C14), and (v) EVAX-Base–NIH-CXR14 (E-B-C14).
Together, these models provide diversity across architecture families and sizes, training datasets, and training paradigms. The pool includes task-specialized models as well as models trained on large combined corpora and foundation-model variants, with different pretraining and supervision setups, to reflect realistic heterogeneity in clinical deployment.
\\
\noindent\textbf{Report Generation:}
We include 19 report-generation models spanning encoder–decoder, autoregressive, and multimodal LLM architectures. These comprise CheXAgent-8B, CheXAgent-2.3B, CheXpert Plus (CheXpert), CheXpert Plus (CheXpert+MIMIC), CheXpert Plus (MIMIC), MAIRA-2, Qwen3-VL-2B, Qwen3-VL-4B, Qwen3-VL-8B, R2GenGPT-Delta, R2GenGPT-Shallow, RaDialog, RadLLaMA-7B, LLaMA-3.2-Vision-11B, LLaVA-1.5-13B, LLaVA-1.5-7B, LLaVA-OneVision-Qwen2-7B, LLaVA-1.6-Mistral-7B, LLaVA-1.6-Vicuna-13B. 
\\
\noindent\textbf{Visual Grounding:} Our 6 grounding models span (i) dedicated phrase-grounding architectures (TransVG/MDETR) with medical pretraining (AGPT), (ii) knowledge-enhanced grounding (AG-KD), and (iii) multimodal radiology foundation models (ChEX, MAIRA-2, RadVLM), providing complementary inductive biases and supervision styles.
\\
\noindent\textbf{Visual Question Answering:} We include 13 diverse multimodal models spanning different model families and scales, including Qwen-based models (Qwen-2B/4B/8B/30B), LLaVA (7B, 13B, OneVision), LLaMA-3.2-Vision, BLIP models trained on CXR and MIMIC, as well as radiology-specific models such as CheXAgent-8B, RaDialog, and NVIDIA CXR NV-Reason-3B.

\begin{figure*}[t]
    \centering
\includegraphics[width=2.1\columnwidth]{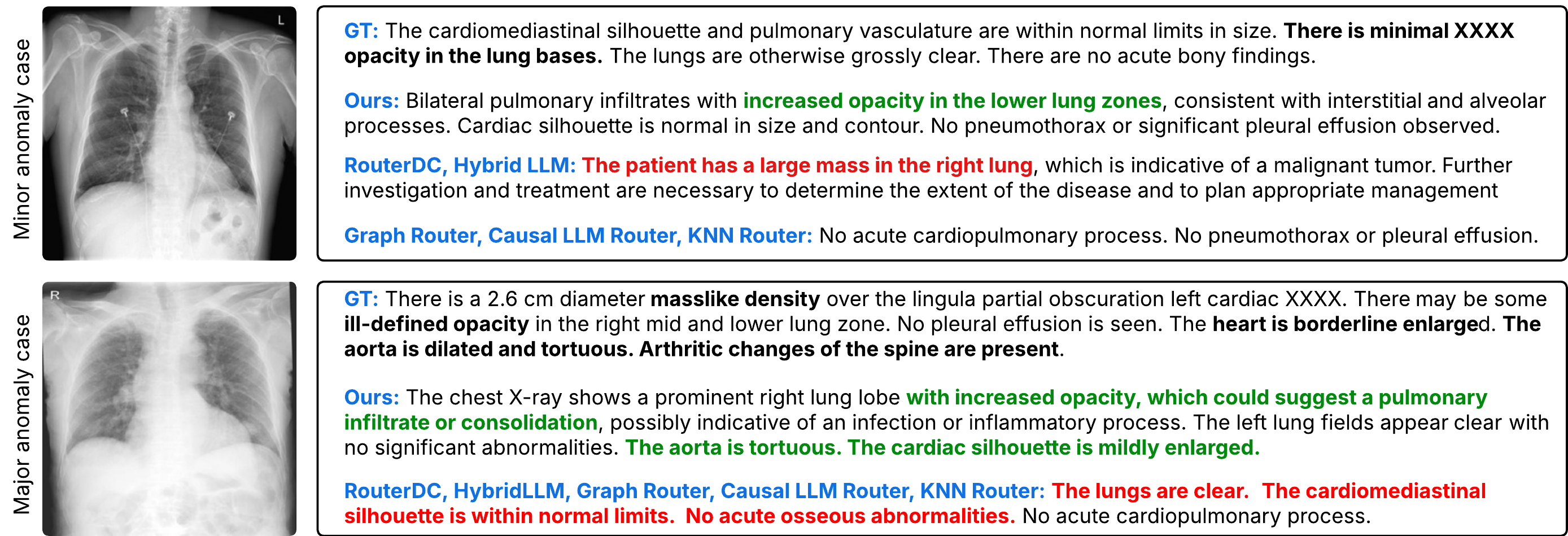}
    \caption{Qualitative report-generation comparison on minor (top) and major (bottom) anomaly cases (GT: ground truth). ToolSelect (ours) routes each query to the appropriate specialist, producing clinically aligned reports that emphasize the correct abnormal regions and severity. Baseline routers often hallucinate severity in minor cases or under-report major findings, yielding generic/misleading text; in the bottom case they all default to the globally strong CheXagent-8B and fail, whereas ToolSelect selects the better-suited CheXpert Plus model (despite lower average performance) and succeeds, highlighting the benefit of query-guided selection.}

\label{fig:11}
\end{figure*}
\subsection{Training and Implementation details}
We use a task-shared ViT-B/16 image encoder $\phi_x$ and CheXbert text encoder $\phi_q$. The router head is a two-layer MLP (hidden width 512) with GELU activations. We optimize the logistic comp-sum surrogate using AdamW (learning rate $3\times10^{-5}$, weight decay $10^{-4}$) for up to 50 epochs with early stopping (patience $=10$, $\text{min\_delta}=10^{-4}$). The batch size is 16 for training, validation, and testing. Regularization includes panel entropy ($\lambda_H=0.05$), an $L_2$ penalty on $r_\theta$, and dropout of 0.1 in the ANP.

\subsection{Baselines:} 
Due to absence of existing baselines for tool selection, we adapt 10 SOTA LLM-routers spanning instance-based, supervised, representation-learning, and structured routing paradigms. We include {KNNRouter} \cite{feng2025fusionfactory} (nearest-neighbour retrieval over historical queries),
{SVMRouter} \cite{llmrouter2025} and {MLPRouter} (supervised classifiers in embedding space), {MFRouter} \cite{ong2024routellm} (matrix
factorization learning latent query/model embeddings), {EloRouter} \cite{ong2024routellm} (that selects the
top-ranked model), {RouterDC} \cite{chen2024routerdc} (dual-contrastive
representation learning), {AutoMix} \cite{madaan2023automix} and {Hybrid LLM} \cite{dinghybrid} (small-to-large model escalation based on
confidence/quality-gap signals), {GraphRouter} \cite{fenggraphrouter} (GNN-based routing over a query--model interaction graph), and
{CausalLLM Router} \cite{ong2024routellm} (routing framed as generation, predicting the selected model name with a finetuned causal LLM).

\section{Results and Analysis}
\subsection{Disease Diagnosis Performance}

\begin{figure*}[t]
    \centering
\includegraphics[width=2.1\columnwidth]{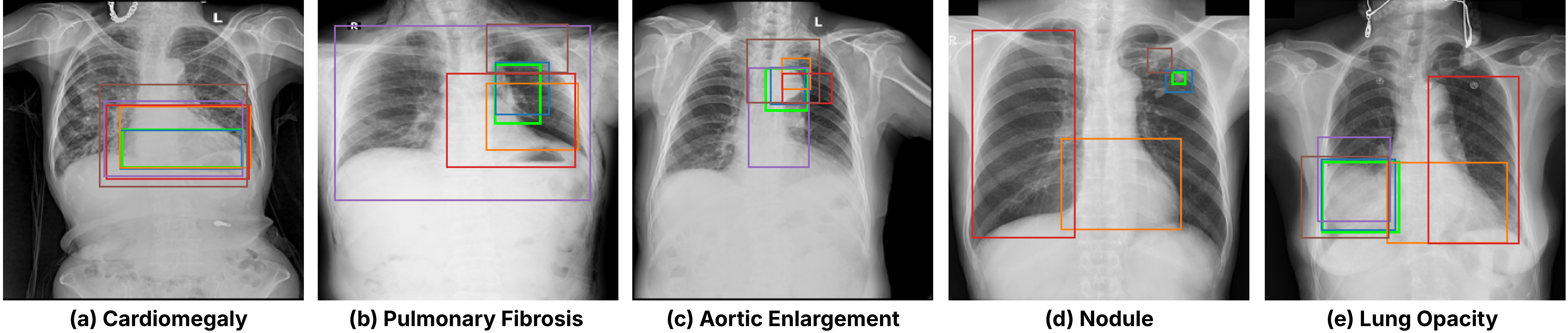}
    \caption{Qualitative phrase-level visual grounding results on chest X-rays. Green denotes the ground-truth box; blue denotes ToolSelect (ours); orange/red/brown/violet denote four baseline routers. ToolSelect consistently routes to a specialist that localizes the queried finding more tightly and accurately, while baselines often miss the target region or return overly coarse boxes that cover irrelevant area.}

\label{fig:11}
\end{figure*}

\begin{table}[t]
\centering
\scriptsize
\setlength{\tabcolsep}{2.2pt}
\caption{Comparison of disease-diagnosis performance (\%) of individual tool candidates and tool selection methods.
Acc: Accuracy; F1: F1 score; P: Precision; R: Recall. (all macro-averaged)
}
\begin{tabular}{lccccc|ccccc}
\hline
\textbf{Method} 
& \multicolumn{5}{c|}{\textbf{Open-I}} 
& \multicolumn{5}{c}{\textbf{VinDr}} \\
\cline{2-6}\cline{7-11}
& \textbf{Acc} & \textbf{F1} & \textbf{P} & \textbf{R} & \textbf{AUC}
& \textbf{Acc} & \textbf{F1} & \textbf{P} & \textbf{R} & \textbf{AUC} \\
\hline
\hline
\multicolumn{11}{c}{\textbf{Tool Candidate Models}} \\
\hline
D-PC 
& 60.22 & 13.35 & 9.72 & 42.39 & 38.74 
& 49.32 & 22.12 & 20.06 & 46.86 & 41.16 \\
D-NIH 
& 53.88 & 12.54 & 9.85 & 42.93 & 36.91 
& 35.50 & 23.90 & 17.86 & 51.70 & 37.21 \\
D-CX 
& 42.02 & 10.59 & 9.12 & 29.71 & 36.24 
& 40.02 & 23.99 & 18.19 & 49.06 & 39.60 \\
D-M-Ch 
& 20.69 & 11.95 & 7.88 & 48.44 & 33.50 
& 26.25 & 24.72 & 16.84 & 62.31 & 36.62 \\
D-M-NB 
& 28.21 & 10.82 & 7.91 & 36.71 & 33.16 
& 26.04 & 24.81 & 16.89 & 62.27 & 37.43 \\
D-RSNA 
& 12.32 & 11.07 & 8.84 & 54.16 & 32.15 
& 19.71 & 24.44 & 16.47 & 64.99 & 33.83 \\
D-All 
& 68.34 & 12.36 & 11.05 & 26.95 & 39.13 
& 63.53 & 24.81 & 21.39 & 43.01 & 44.60 \\
D-MV2 
& 78.10 & 9.61 & 37.90 & 25.46 & 38.27 
& 68.45 & 9.21 & 45.65 & 13.68 & 43.96 \\
R-MGCA 
& 77.96 & 8.61 & 27.39 & 24.29 & 37.46 
& 68.68 & 10.69 & 44.37 & 14.56 & 46.71 \\
R-MKP 
& 78.17 & 8.21 & 20.73 & 23.83 & 37.73 
& 68.77 & 10.83 & 51.60 & 14.51 & 45.61 \\
R-BioViL 
& 78.13 & 6.82 & 30.45 & 22.55 & 37.69 
& 68.57 & 9.60 & 61.65 & 13.68 & 44.65 \\
R-All 
& 84.08 & 6.61 & 47.12 & 14.49 & 40.84 
& 75.66 & 4.70 & 47.45 & 6.87 & 40.71 \\
E-T-CX
& 28.63 & 9.12 & 9.54 & 37.83 & 35.67 
& 34.21 & 20.82 & 21.48 & 47.24 & 39.74 \\
E-S-CX 
& 28.53 & 11.44 & 11.50 & 39.91 & 35.43 
& 33.73 & 21.66 & 27.41 & 48.64 & 39.88 \\
E-T-C14 
& 78.21 & 7.37 & 31.84 & 22.90 & 37.19 
& 68.72 & 10.21 & 53.87 & 14.32 & 47.98 \\
E-S-C14 
& 78.03 & 8.01 & 31.92 & 23.46 & 36.55 
& 68.85 & 12.50 & 58.65 & 15.81 & 48.22 \\
E-B-C14 
& 78.13 & 7.97 & 21.20 & 23.35 & 37.76 
& 68.92 & 13.06 & 48.91 & 16.09 & 48.24 \\
\hline
\multicolumn{11}{c}{\textbf{Tool Selection Methods}} \\
\hline
Random              & 60.15    & 9.54    & 7.56    & 27.19    & 32.20    & 53.32 & 20.81 & 18.19 & 32.07 & 39.01\\
AutoMix     & 51.14 & 10.53 & 16.07 & 23.56 & 35.68 & 51.67 & 20.96 & 18.28 & 33.32 & 37.06 \\
SVM Router
& 33.33 & 11.76 & 13.70 & 37.59 & 36.77
& 72.58 & 13.18 & 27.37 & 14.01 & 41.32 \\

Graph Router
& 12.32 & 11.07 & 8.84  & 54.16 & 32.15
& 68.71 & 10.24 & 53.24 & 14.34 & 46.00 \\

MF Router
& 33.97 & 11.98 & 13.41 & 38.41 & 36.34
& 74.24 & 9.59  & 27.23 & 10.32 & 40.71 \\

RouterDC
& 81.52 & 8.51  & 33.24 & 19.47 & 38.11
& 66.15 & 16.66 & 27.20 & 20.95 & 46.52 \\

Hybrid LLM
& 66.13 & 8.00  & 13.80 & 19.06 & 35.58
& 67.21 & 18.27 & 27.09 & 20.32 & 46.66 \\

KNN Router
& 30.16 & 11.52 & 11.63 & 38.75 & 35.09
& 74.23 & 8.92  & 26.04 & 9.83  & 41.18 \\

Elo Router
& 28.53 & 11.44 & 11.50 & 39.91 & 35.43
& 75.66 & 4.70  & 47.45 & 6.87  & 40.71 \\

MLP Router
& 33.69 & 9.01  & 11.46 & 31.92 & 34.29
& 73.33 & 12.24 & 29.12 & 12.59 & 40.83 \\

Causal LLM
& 12.32 & 11.07 & 8.84  & 54.16 & 32.15
& 19.71 & 24.44 & 16.47 & 64.99 & 33.83 \\

ToolSelect   &  \textbf{87.64}   & \textbf{43.80}    &  \textbf{48.55}   & \textbf{57.31}    & \textbf{45.09}    & \textbf{83.39} & \textbf{59.88} & \textbf{68.62} & \textbf{65.15} & \textbf{49.91} \\ \hline
Oracle              & 99.22 & 58.00 & 94.80 & 61.11 & 60.71 &  98.45 & 66.22 & 93.58 & 66.67 & 66.54 \\
\hline
\end{tabular}
\label{tab:diagnosis_openi_vindr}
\end{table}

\begin{table}[t]
\centering
\scriptsize
\setlength{\tabcolsep}{6pt}
\caption{Report-generation performance across models (in \%). 
C: CheXpert; M: MIMIC; C+M: CheXpert+MIMIC.
R-L: ROUGE-L; MR: METEOR; F1-RG: F1-RadGraph; SemB: SemBScore; Rate: RateScore. Higher is better.}
\begin{tabular}{lcccccc}
\toprule
\textbf{Model} & \textbf{SemB} & \textbf{F1-RG} & \textbf{R-L} & \textbf{MR} & \textbf{Rate} \\
\cmidrule(lr){1-6}
\multicolumn{6}{c}{\textbf{Tool Candidate Models}} \\
\cmidrule(lr){1-6}
CheXAgent-8B & 41.91 & 16.33 & 21.97 & 25.00 & 55.06 \\
CheXagent-2-3B & 34.44 & 11.47 & 15.26 & 23.64 & 54.24 \\
CheXpert Plus (C) & 42.86 & 15.71 & 19.48 & 30.55 & 56.59 \\
CheXpert Plus (C+M) & 41.81 & 16.10 & 20.35 & 30.88 & 57.05 \\
CheXpert Plus (M) & 43.32 & 10.71 & 19.91 & 28.99 & 54.31 \\
MAIRA-2 & 36.90 & 9.56 & 14.34 & 16.44 & 45.55 \\
Qwen3-VL-2B & 38.62 & 12.32 & 16.65 & 26.73 & 51.72 \\
Qwen3-VL-4B & 45.38 & 10.63 & 16.35 & 22.52 & 51.33 \\
Qwen3-VL-8B & 37.00 & 10.62 & 15.47 & 26.77 & 50.65 \\
R2GenGPT-Delta & 43.30 & 14.07 & 20.25 & 29.78 & 53.68 \\
R2GenGPT-Shallow & 40.41 & 15.56 & 20.37 & 30.73 & 52.95 \\
RaDialog & 35.28 & 11.04 & 20.13 & 25.22 & 50.31 \\
RadLLaMA-7B & 35.47 & 13.63 & 11.40 & 24.33 & 49.98 \\
LLaMA-3.2-Vision-11B & 26.71 & 7.00 & 13.49 & 20.15 & 47.26 \\
LLaVA-1.5-13B & 23.72 & 1.57 & 11.66 & 10.89 & 39.57 \\
LLaVA-1.5-7B & 20.37 & 3.05 & 12.10 & 12.28 & 40.09 \\
LLaVA-OneVision-Qwen2-7B & 25.84 & 9.37 & 15.32 & 22.19 & 48.78 \\
LLaVA-1.6-Mistral-7B & 26.84 & 8.53 & 13.08 & 19.97 & 46.85 \\
LLaVA-1.6-Vicuna-13B & 24.83 & 8.03 & 15.37 & 20.26 & 46.31 \\
\cmidrule(lr){1-6}
\multicolumn{6}{c}{\textbf{Tool Selection Methods}} \\
\cmidrule(lr){1-6}
Random        & 36.04 & 11.71 & 16.42 & 24.23 & 50.56 \\
AutoMix       & 23.72 & 1.57  & 11.66 & 10.89 & 39.57 \\
SVM Router   & 42.39 & 16.87 & 21.97 & 26.04 & 55.88 \\
Graph Router & 41.91 & 16.33 & 21.97 & 25.00 & 55.06 \\
MF Router    & 46.53 & 22.40 & 21.77 & 31.25 & 56.76 \\
RouterDC      & 41.89 & 16.01 & 20.27 & 30.61 & 56.94 \\
Hybrid LLM   & 44.13 & 16.77 & 22.10 & 24.99 & 55.60 \\
KNN Router   & 42.64 & 15.74 & 19.83 & 30.29 & 56.36 \\
Elo Router   & 41.91 & 16.33 & 21.97 & 25.00 & 55.06 \\
MLP Router   & 42.39 & 16.87 & 21.97 & 26.04 & 55.88 \\
Causal LLM Router & 42.59 & 15.74 & 19.54 & 30.39 & 56.83 \\
ToolSelect (Ours)   & \textbf{47.92} & \textbf{24.17} & \textbf{22.44} & \textbf{32.51} & \textbf{57.86} \\ \hline
Oracle        & 50.17 & 27.70 & 23.85 & 34.31 & 59.60 \\
\bottomrule
\end{tabular}
\label{tab:report_generation_metrics_only}
\end{table}

\begin{table*}[t]
\centering
\scriptsize
\setlength{\tabcolsep}{1pt}
\caption{Per-phrase performance (in \%) with grounding metrics appended (mAP and Mean IoU).}
\scalebox{1.05}{%
\begin{tabular}{lccccccccccc|ccc}
\hline
\textbf{Model} &
\textbf{Aortic enl.} & \textbf{Atel.} & \textbf{Calc.} & \textbf{Cardio.} &
\textbf{Cons.} & \textbf{ILD} & \textbf{Infil.} & \textbf{Lung Opac.} &
\textbf{Nodule/Mass} & \textbf{Other les.} & \textbf{Pulm. fib.} &
\textbf{mAP@0.25} & \textbf{mAP@0.5} & \textbf{Mean IoU} \\
\hline
\multicolumn{14}{c}{\textbf{Tool Candidate Models}} \\
\hline
AG-KD         & 66.25 & 33.41 & 17.33 & 70.12 & 39.45 & 22.90 & 31.48 & 28.71 & 7.58  & 12.70 & 22.55 & 45.84 & 32.96 & 40.66 \\
MDETR-AGPT    & 25.84 & 6.80  & 1.06  & 40.52 & 6.19  & 14.11 & 5.07  & 5.18  & 2.91  & 6.29  & 3.10  & 18.96 & 1.76  & 16.52 \\
TransVG-AGPT  & 14.37 & 5.98  & 1.74  & 41.52 & 13.12 & 13.24 & 11.57 & 5.92  & 2.23  & 4.08  & 1.73  & 16.83 & 3.22  & 14.06 \\
ChEX          & 53.00 & 49.44 & 18.86 & 71.50 & 54.31 & 14.21 & 35.99 & 32.65 & 13.71 & 27.72 & 20.22 & 59.69 & 39.42 & 40.25 \\
MAIRA-2       & 16.34 & 13.19 & 6.11  & 38.32 & 14.00 & 17.18 & 13.18 & 11.32 & 1.81  & 3.04  & 5.81  & 24.18 & 4.60  & 16.67 \\
RadVLM        & 18.99 & 29.90 & 10.24 & 52.96 & 25.02 & 23.78 & 19.71 & 16.92 & 5.52  & 11.32 & 14.29 & 39.51 & 18.45 & 23.14 \\
\hline
\multicolumn{14}{c}{\textbf{Tool Selection Methods}} \\
\hline
Random              & 37.76 & 30.37 & 9.21  & 56.78 & 34.67 & 25.50 & 27.14 & 22.68 & 7.77  & 20.45 & 13.88 & 50.90 & 10.12 & 32.41 \\
AutoMix     & 65.73 & 40.25 & 5.59  & 69.05 & 45.80 & 29.61 & 39.71 & 31.54 & 8.01  & 19.97 & 18.57 & 54.13 & 35.21 & 45.94 \\
SVM Router         & 27.58 & 5.44  & 1.61  & 42.36 & 6.95  & 12.32 & 4.34  & 2.85  & 3.06  & 7.72  & 4.33  & 18.44 & 1.87  & 20.42 \\
Elo Router         & 27.61 & 5.44  & 1.61  & 42.36 & 6.95  & 12.32 & 4.34  & 2.85  & 3.06  & 7.72  & 4.33  & 18.44 & 1.87  & 20.43 \\
Graph Router       & 64.03 & 40.25 & 2.76  & 70.13 & 45.80 & 30.14 & 39.71 & 31.57 & 9.08  & 22.90 & 19.13 & 55.17 & 35.07 & 45.92 \\
MF Router          & 64.91 & 40.25 & 5.59  & 68.75 & 45.80 & 30.22 & 39.71 & 32.19 & 8.01  & 17.29 & 17.66 & 53.79 & 34.83 & 45.42 \\
RouterDC            & 51.16 & 39.55 & 3.11  & 65.13 & 30.71 & 28.69 & 37.50 & 21.30 & 8.05  & 5.89  & 15.20 & 46.25 & 27.12 & 38.79 \\
Hybrid LLM & 59.88 & 39.70 & 2.78  & 50.17 & 33.97 & 31.79 & 37.11 & 28.34 & 5.27  & 16.10 & 14.75 & 45.49 & 30.20 & 38.38 \\
KNN Router         & 27.42 & 5.44  & 1.61  & 42.36 & 6.95  & 12.32 & 4.34  & 2.85  & 3.06  & 7.72  & 4.33  & 18.37 & 1.87  & 20.37 \\
Causal LLM Router & 65.73 & 40.25 & 5.59 & 69.05
& 45.80 & 29.61 & 39.71 & 31.54
& 8.01 & 19.97 & 18.57
& 54.12 & 35.21 & 45.94 \\
MLP Router         & 27.58 & 5.44  & 1.61  & 42.19 & 6.95  & 12.32 & 4.34  & 2.85  & 3.06  & 7.72  & 4.33  & 18.44 & 1.87  & 20.38 \\
ToolSelect (Ours)   & \textbf{67.16}    & \textbf{52.06}    & \textbf{19.72}    & \textbf{72.60}    & \textbf{59.56}    & \textbf{34.23}    & \textbf{41.72}    & \textbf{36.61}    & \textbf{15.82}    & \textbf{34.25}    & \textbf{26.81}    & \textbf{63.98}    & \textbf{48.39}    & \textbf{50.08}    \\ \hline
Oracle              & 69.29 & 63.19 & 23.92 & 76.77 & 63.68 & 49.34 & 58.74 & 49.01 & 19.06 & 47.33 & 32.63 & 78.34 & 56.31 & 56.20 \\
\hline
\end{tabular}%
}
\label{tab:phrase_transposed_with_map}
\end{table*}
\begin{table*}[t]
\centering
\scriptsize
\setlength{\tabcolsep}{3pt}
\caption{Comparison across VQA task (\%). Top row: Tool candidates (Q30B:Qwen-30B; LOV:LLaVA-OneVision; Q4B:Qwen-4B; Q8B:Qwen-8B; L7B:LLaVA-7B; L13B:LLaVA-13B; Q2B:Qwen-2B; L3.2V:LLaMA-3.2-Vision; BCXR:BLIP-CXR; BMIM:BLIP-MIMIC; CXA:CheXagent-8B; RDlg:RaDialog; NV: NVIDIA CXR NV-Reason-3B). Bottom row: Selection methods.}
\begin{tabular}{lcccccccccccccc}
\hline
\textbf{Tool Candidates} 
& \textbf{Q30B} 
& \textbf{LOV} 
& \textbf{Q4B} 
& \textbf{Q8B} 
& \textbf{L7B} 
& \textbf{L13B} 
& \textbf{Q2B} 
& \textbf{L3.2V} 
& \textbf{BCXR} 
& \textbf{BMIM} 
& \textbf{CXA} 
& \textbf{RDlg} 
& \textbf{NV} \\
\hline
\textbf{VQA Acc (\%)} 
& 60.63 & 58.91 & 57.76 & 56.32 & 54.60 & 53.74 & 53.16 & 33.05 & 25.00 & 23.60 & 47.41 & 38.51 & 63.22 \\
\hline
\textbf{Selection Methods} 
& \textbf{Random} 
& \textbf{KNN} 
& \textbf{SVM} 
& \textbf{MLP} 
& \textbf{MF} 
& \textbf{Elo} 
& \textbf{Graph}
& \textbf{RouterDC} 
& \textbf{AutoMix} 
& \textbf{Hybrid LLM}  
& \textbf{Causal LLM} 
& \textbf{ToolSelect} 
& \textbf{Oracle} \\
\hline
\textbf{VQA Acc (\%)} 
& 46.49 & 41.52 & 47.37 & 47.08 & 36.55 & 63.74 & 47.37 & 45.61 & 24.27 & 42.69 & 47.37 &\textbf{ 72.01} & 96.49 \\
\hline
\end{tabular}
\label{tab:vqa_router_vs_model}
\end{table*}
Table 1 shows that specialist models occupy markedly different points on the precision-recall trade-off, and these rankings shift substantially between Open-I and VinDr.

\noindent\textbf{Failure modes and complementarity of specialist models}: Table~1 shows that no single specialist is uniformly reliable under dataset shift, and conclusions depend strongly on the metric. For example, R-All attains the highest accuracy on both Open-I ($84.08\%$) and VinDr ($75.66\%$) but has the lowest F1 (Open-I $6.61\%$, VinDr $4.70\%$) and very low recall (Open-I $14.49\%$, VinDr $6.87\%$), reflecting a conservative regime that misses positives. In contrast, D-PC/D-NIH achieve higher Open-I recall ($42.39\%/42.93\%$) at low precision ($\sim$10\%), indicating an over-calling regime, while MIMIC-trained models exhibit shift-induced miscalibration (e.g., D-M-Ch: recall $48.44\%$ but Open-I accuracy $20.69\%$). The ``best'' model is context-dependent: candidates trade precision for recall differently across Open-I and VinDr (e.g., D-RSNA reaches $64.99\%$ recall on VinDr but degrades sharply on Open-I). This matters clinically: high-recall specialists may be preferred for screening/triage (e.g., D-RSNA $64.99\%$, D-M-Ch/D-M-NB $62.31\%/62.27\%$ recall on VinDr), whereas high-precision specialists may be preferred for prioritization/confirmation (e.g., R-BioViL $61.65\%$, E-S-C14 $58.65\%$ precision on VinDr). 

\noindent\textbf{Random vs.\ single specialists vs.\ oracle.}
Random routing is a lower bound when selection ignores query-specific information (Open-I: $9.54\%$ F1/$60.15\%$ accuracy; VinDr: $20.81\%$ F1/$53.32\%$ accuracy). The Oracle attains $58.00\%$ F1 on Open-I and $66.22\%$ on VinDr, indicating large headroom from per-instance specialization. On VinDr, the best single-model F1 is only $24.81\%$ (D-All/D-M-NB), close to Random, while Oracle is much higher, showing that the optimal specialist varies substantially across cases.

\noindent\textbf{Baseline routers:}
Classical and heuristic routers give mixed improvements and can trail strong single tools. On Open-I, the best baseline F1 is MF Router at $11.98\%$ (+$2.44$ over Random), while some achieve high accuracy but low F1 (e.g., RouterDC $81.52\%$ accuracy vs $8.51\%$ F1), consistent with conservative selection that misses positives. On VinDr, the best baseline F1 is Causal LLM at $24.44\%$, close to the best single-model F1 ($24.81\%$) and far below Oracle ($66.22\%$), indicating unreliable per-instance selection.

\noindent\textbf{ToolSelect closes the Random--Oracle gap.}
ToolSelect consistently outperforms all baselines and captures a large fraction of the Oracle gap. On Open-I, it achieves $43.80\%$ F1/$87.64\%$ accuracy, improving over the best baseline by $+31.82$ F1 points (vs $11.98\%$) and over the best single-model F1 by $+30.45$ (vs $13.35\%$). On VinDr, it achieves $59.88\%$ F1/$83.39\%$ accuracy, improving over the best baseline by $+35.44$ F1 points (vs $24.44\%$) and over the best single-model F1 by $+35.07$ (vs $24.81\%$).

\subsection{Report Generation}
Consistent with diagnosis (\S4.1), \Cref{tab:report_generation_metrics_only} shows large variability across report-generation tools and modest absolute performance. Radiology-tuned generators achieve the highest clinical correctness (CheXAgent-8B: $16.33\%$ F1-RadGraph; CheXpert Plus (C+M): $16.10\%$), while several general-purpose VLMs degrade sharply (e.g., LLaVA-1.5-13B: $1.57\%$; LLaVA-1.5-7B: $3.05\%$), indicating that these models individually do not ensure clinical faithfulness.
\\
\noindent\textbf{Random vs.\ single vs.\ Oracle.} Random selection yields limited correctness (F1-RadGraph $11.71\%$, RateScore $50.56$). The best single generator reaches $16.33\%$ F1-RadGraph (CheXAgent-8B), whereas Oracle attains $27.70\%$ and $59.60$ RateScore, revealing substantial per-instance heterogeneity and a large Oracle gap.
\\
\noindent\textbf{Baseline selectors.} Baseline routers improve over Random, with MF Router at $22.40\%$ F1-RadGraph and RouterDC at $16.01\%$, but remain well below Oracle ($27.70\%$), suggesting limited ability to identify the best specialist per query.
\\
\noindent\textbf{ToolSelect.} ToolSelect achieves the best routing performance ($24.17\%$ F1-RadGraph, $57.86$ RateScore), improving over Random by $+12.46$ points and over the best single generator by $+7.84$, and surpassing the strongest baseline (MF Router) by $+1.77$ F1-RadGraph. Qualitative examples in Fig.~5 show that ToolSelect selects specialists that produce more clinically aligned reports than baseline routing. See Suppl. E.1 for \textbf{ablation on reference point / aggregation}. 
\subsection{Visual Grounding}
\noindent\textbf{Analysis of complementarity.}
\Cref{tab:phrase_transposed_with_map} shows grounding is highly phrase-dependent: different specialists excel on different findings and no single candidate dominates. ChEX is the most consistently strong single tool (e.g., best on Atelectasis $49.44$, Calcification $18.86$, Cardiomegaly $71.50$), suggesting it is effective for well-localized findings with consistent appearance and clear spatial supervision. However, clinically important crossovers remain: AG-KD is stronger on Aortic enlargement and Pulmonary fibrosis ($22.55$ vs $20.22$), indicating better sensitivity to global morphology and diffuse chronic patterns, while RadVLM is best on ILD, suggesting stronger language--vision alignment for heterogeneous interstitial disease. Overall, the best tool depends on the image and phrase, not only on average performance.
\\
\noindent\textbf{Hard phrases and weak specialists.}
Some phrases remain challenging for all tools (e.g., Nodule/Mass peaks at only $13.71$; ``Other lesion'' is modest across candidates), consistent with subtle lesions, IoU sensitivity to small mis-localizations, and annotation variability. We also observe weak specialists: MDETR-AGPT and TransVG-AGPT have low mean IoU ($16.52$ and $14.06$) and near-zero performance on several phrases (e.g., Calcification, Nodule/Mass). 
\\
\noindent\textbf{Random vs.\ specialists vs.\ Oracle and ToolSelect.}
Random routing is non-trivial (IoU $32.41$) due to strong candidates (AG-KD/ChEX), but remains far from Oracle ($56.20$), and the gap to the best single model ($40.66$) indicates strong per-instance heterogeneity. As two specialists are already strong, many baseline selectors yield limited gains over choosing a strong fixed tool and weak routing can even hurt. ToolSelect nonetheless closes much of the Random--Oracle gap (mean IoU $50.08$) and improves over the best individual grounding model by $\approx 10$ IoU points, reaching mAP@0.5 $48.39$ (about $+9$ over the best single), demonstrating effective query-conditioned matching across phrases and images. See Fig. 6 for qualitative performance evaluation.

\subsection{Visual Question Answering}
Table~4 shows wide variation across VQA tools, with accuracy ranging from $23.60\%$ (BLIP-MIMIC) to $63.22\%$ (NV), and a tight cluster of Qwen and LLaVA variants around the $50\%$ range. This suggests several candidates are similarly strong on average, motivating query-conditioned selection when the best tool varies by question.
\\
\textbf{Do tool selection methods help?} ToolSelect improves beyond the best individual model, reaching $72.01\%$ accuracy compared to the strongest single model at $63.22\%$. Random selection is poor ($46.49\%$), far below the best individual tool and far below Oracle ($96.49\%$). This very large Oracle gap indicates strong per-instance specialization.
Importantly, many classical routers underperform strong single models (e.g., KNN $41.52\%$, Graph $47.37\%$), highlighting that learning to route is non-trivial and that weak routing can destroy the benefit of having strong specialists. The fact that ToolSelect closes a meaningful portion of this gap (to $72.01\%$) suggests it extracts predictive signals from the query and tool behavior to choose the correct specialist more often than naive or heuristic routing.
\section{Conclusion}
In this paper, we studied query-conditioned selection among heterogeneous task-specialized models as tools for agentic healthcare. To address this, we introduced \textbf{ToolSelect}, an Attentive Neural Process-based selector that conditions on the incoming query and per-tool behavioral summaries to route each request to an appropriate specialist model while keeping all tool models frozen.
\\
To enable systematic evaluation of tool candidate selection in a clinically grounded setting, we introduced an agentic chest X-ray environment with a diverse suite of specialist tools spanning disease diagnosis, report generation, visual grounding, and VQA, and developed \textbf{ToolSelectBench}. Across these task families, our results  show substantial complementarity among individual specialist models and consistently demonstrate that our learned selection mechanism outperforms SOTA routing baselines. Overall, our findings indicate that reliable performance in multimodal clinical agentic pipelines depends not only on usage of single specialized tool but combination and selection of multiple models via query-conditioned tool selection mechanism. 
\section*{Impact Statement}
This paper studies tool selection in agentic machine learning systems, motivated by healthcare applications where multiple task-specialized models must be orchestrated reliably. Our primary goal is to advance the methodological foundations of model and tool selection by formalizing query-conditioned routing among heterogeneous candidates and demonstrating its effectiveness empirically.
Potential positive impacts include improved reliability, efficiency, and transparency in agentic systems that integrate multiple pretrained models, particularly in medical imaging and clinical decision-support workflows. By enabling agents to select appropriate specialist tools rather than invoking all available models, the proposed approach may reduce computational cost and support more structured, auditable system design.
At the same time, our work does not propose autonomous clinical decision-making, and all medical use cases considered are intended to support, rather than replace, human expertise. As with any machine learning system applied in healthcare, improper deployment without clinical oversight could pose risks. These concerns are not unique to our method and reflect broader challenges in the safe integration of AI systems into medical practice.
Overall, this work contributes to the development of agentic architectures in machine learning. We do not anticipate new ethical risks beyond those already well studied in the deployment of machine learning systems, particularly in healthcare, but emphasize that responsible use requires careful validation, transparency, and human-in-the-loop oversight.
\bibliography{example_paper}

@article{thirunavukarasu2023large,
  title={Large language models in medicine},
  author={Thirunavukarasu, Arun James and Ting, Darren Shu Jeng and Elangovan, Kabilan and Gutierrez, Laura and Tan, Ting Fang and Ting, Daniel Shu Wei},
  journal={Nature medicine},
  volume={29},
  number={8},
  pages={1930--1940},
  year={2023},
  publisher={Nature Publishing Group US New York}
}

@article{naveed2025comprehensive,
  title={A comprehensive overview of large language models},
  author={Naveed, Humza and Khan, Asad Ullah and Qiu, Shi and Saqib, Muhammad and Anwar, Saeed and Usman, Muhammad and Akhtar, Naveed and Barnes, Nick and Mian, Ajmal},
  journal={ACM Transactions on Intelligent Systems and Technology},
  volume={16},
  number={5},
  pages={1--72},
  year={2025},
  publisher={ACM New York, NY}
}

@article{wang2024survey,
  title={A survey on large language model based autonomous agents},
  author={Wang, Lei and Ma, Chen and Feng, Xueyang and Zhang, Zeyu and Yang, Hao and Zhang, Jingsen and Chen, Zhiyuan and Tang, Jiakai and Chen, Xu and Lin, Yankai and others},
  journal={Frontiers of Computer Science},
  volume={18},
  number={6},
  pages={186345},
  year={2024},
  publisher={Springer}
}

@article{guo2024large,
  title={Large language model based multi-agents: A survey of progress and challenges},
  author={Guo, Taicheng and Chen, Xiuying and Wang, Yaqi and Chang, Ruidi and Pei, Shichao and Chawla, Nitesh V and Wiest, Olaf and Zhang, Xiangliang},
  journal={arXiv preprint arXiv:2402.01680},
  year={2024}
}

@article{qiu2024llm,
  title={LLM-based agentic systems in medicine and healthcare},
  author={Qiu, Jianing and Lam, Kyle and Li, Guohao and Acharya, Amish and Wong, Tien Yin and Darzi, Ara and Yuan, Wu and Topol, Eric J},
  journal={Nature Machine Intelligence},
  volume={6},
  number={12},
  pages={1418--1420},
  year={2024},
  publisher={Nature Publishing Group UK London}
}

@article{wang2025survey,
  title={A survey of llm-based agents in medicine: How far are we from baymax?},
  author={Wang, Wenxuan and Ma, Zizhan and Wang, Zheng and Wu, Chenghan and Ji, Jiaming and Chen, Wenting and Li, Xiang and Yuan, Yixuan},
  journal={arXiv preprint arXiv:2502.11211},
  year={2025}
}

@incollection{mofrad2025limitations,
  title={Limitations of large language models in healthcare systems},
  author={Mofrad, Farshid Babapour and Yousefzamani, Midya},
  booktitle={Applications of Large Language Models (LLM) in Healthcare Systems},
  pages={234--262},
  year={2025},
  publisher={Chapman and Hall/CRC}
}

@article{fallahpour2025medrax,
  title={Medrax: Medical reasoning agent for chest x-ray},
  author={Fallahpour, Adibvafa and Ma, Jun and Munim, Alif and Lyu, Hongwei and Wang, Bo},
  journal={arXiv preprint arXiv:2502.02673},
  year={2025}
}

@article{tang2025endoagent,
  title={EndoAgent: A Memory-Guided Reflective Agent for Intelligent Endoscopic Vision-to-Decision Reasoning},
  author={Tang, Yi and Wang, Kaini and Chen, Yang and Zhou, Guangquan},
  journal={arXiv preprint arXiv:2508.07292},
  year={2025}
}

@article{huang2026surgical,
  title={Surgical AI Copilot: Energy-Based Fourier Gradient Low-Rank Adaptation for Surgical LLM Agent Reasoning and Planning},
  author={Huang, Jiayuan and He, Runlong and Khan, Danyal Zaman and Mazomenos, Evangelos B and Stoyanov, Danail and Marcus, Hani and Jiang, Linzhe and Clarkson, Matthew J and Hoque, Mobarak I},
  year={2026}
}

@inproceedings{sanner2021reliable,
  title={How reliable are out-of-distribution generalization methods for medical image segmentation?},
  author={Sanner, Antoine and Gonzalez, Camila and Mukhopadhyay, Anirban},
  booktitle={DAGM German Conference on Pattern Recognition},
  pages={604--617},
  year={2021},
  organization={Springer}
}

@article{maleki2022generalizability,
  title={Generalizability of machine learning models: quantitative evaluation of three methodological pitfalls},
  author={Maleki, Farhad and Ovens, Katie and Gupta, Rajiv and Reinhold, Caroline and Spatz, Alan and Forghani, Reza},
  journal={Radiology: Artificial Intelligence},
  volume={5},
  number={1},
  pages={e220028},
  year={2022},
  publisher={Radiological Society of North America}
}

@article{musa2025systematic,
  title={A Systematic Review of Cross-Population Shifts in Medical Imaging Analysis with Deep Learning},
  author={Musa, Aminu and Prasad, Rajesh and Onwualu, Peter and Hernandez, Monica},
  year={2025}
}

@inproceedings{mao2023cross,
  title={Cross-entropy loss functions: Theoretical analysis and applications},
  author={Mao, Anqi and Mohri, Mehryar and Zhong, Yutao},
  booktitle={International conference on Machine learning},
  pages={23803--23828},
  year={2023},
  organization={pmlr}
}

@article{mao2023two,
  title={Two-stage learning to defer with multiple experts},
  author={Mao, Anqi and Mohri, Christopher and Mohri, Mehryar and Zhong, Yutao},
  journal={Advances in neural information processing systems},
  volume={36},
  pages={3578--3606},
  year={2023}
}

@article{kim2019attentive,
  title={Attentive neural processes},
  author={Kim, Hyunjik and Mnih, Andriy and Schwarz, Jonathan and Garnelo, Marta and Eslami, Ali and Rosenbaum, Dan and Vinyals, Oriol and Teh, Yee Whye},
  journal={arXiv preprint arXiv:1901.05761},
  year={2019}
}

@article{jaeger2014two,
  title={Two public chest X-ray datasets for computer-aided screening of pulmonary diseases},
  author={Jaeger, Stefan and Candemir, Sema and Antani, Sameer and W{\'a}ng, Y{\`\i}-Xi{\'a}ng J and Lu, Pu-Xuan and Thoma, George},
  journal={Quantitative imaging in medicine and surgery},
  volume={4},
  number={6},
  pages={475},
  year={2014}
}

@article{nguyen2022vindr,
  title={VinDr-CXR: An open dataset of chest X-rays with radiologist’s annotations},
  author={Nguyen, Ha Q and Lam, Khanh and Le, Linh T and Pham, Hieu H and Tran, Dat Q and Nguyen, Dung B and Le, Dung D and Pham, Chi M and Tong, Hang TT and Dinh, Diep H and others},
  journal={Scientific Data},
  volume={9},
  number={1},
  pages={429},
  year={2022},
  publisher={Nature Publishing Group UK London}
}

@article{pal2025rexvqa,
  title={ReXVQA: A Large-scale Visual Question Answering Benchmark for Generalist Chest X-ray Understanding},
  author={Pal, Ankit and Lee, Jung-Oh and Zhang, Xiaoman and Sankarasubbu, Malaikannan and Roh, Seunghyeon and Kim, Won Jung and Lee, Meesun and Rajpurkar, Pranav},
  journal={arXiv preprint arXiv:2506.04353},
  year={2025}
}

@inproceedings{cohen2022torchxrayvision,
  title={TorchXRayVision: A library of chest X-ray datasets and models},
  author={Cohen, Joseph Paul and Viviano, Joseph D and Bertin, Paul and Morrison, Paul and Torabian, Parsa and Guarrera, Matteo and Lungren, Matthew P and Chaudhari, Akshay and Brooks, Rupert and Hashir, Mohammad and others},
  booktitle={International Conference on Medical Imaging with Deep Learning},
  pages={231--249},
  year={2022},
  organization={PMLR}
}

@article{yao2025eva,
  title={Eva-x: A foundation model for general chest x-ray analysis with self-supervised learning},
  author={Yao, Jingfeng and Wang, Xinggang and Song, Yuehao and Zhao, Huangxuan and Ma, Jun and Chen, Yajie and Liu, Wenyu and Wang, Bo},
  journal={npj Digital Medicine},
  volume={8},
  number={1},
  pages={678},
  year={2025},
  publisher={Nature Publishing Group UK London}
}

@article{feng2025fusionfactory,
  title={FusionFactory: Fusing LLM Capabilities with Multi-LLM Log Data},
  author={Feng, Tao and Zhang, Haozhen and Lei, Zijie and Han, Pengrui and Patwary, Mostofa and Shoeybi, Mohammad and Catanzaro, Bryan and You, Jiaxuan},
  journal={arXiv preprint arXiv:2507.10540},
  year={2025}
}

@article{chen2024routerdc,
  title={Routerdc: Query-based router by dual contrastive learning for assembling large language models},
  author={Chen, Shuhao and Jiang, Weisen and Lin, Baijiong and Kwok, James and Zhang, Yu},
  journal={Advances in Neural Information Processing Systems},
  volume={37},
  pages={66305--66328},
  year={2024}
}

@article{madaan2023automix,
  title={AutoMix: Automatically Mixing Language Models},
  author={Madaan, Aman and Aggarwal, Pranjal and Anand, Ankit and Potharaju, Srividya Pranavi and Mishra, Swaroop and Zhou, Pei and Gupta, Aditya and Rajagopal, Dheeraj and Kappaganthu, Karthik and Yang, Yiming and others},
  journal={CoRR},
  year={2023}
}

@inproceedings{dinghybrid,
  title={Hybrid LLM: Cost-Efficient and Quality-Aware Query Routing},
  author={Ding, Dujian and Mallick, Ankur and Wang, Chi and Sim, Robert and Mukherjee, Subhabrata and R{\"u}hle, Victor and Lakshmanan, Laks VS and Awadallah, Ahmed Hassan},
  booktitle={The Twelfth International Conference on Learning Representations}
}

@inproceedings{fenggraphrouter,
  title={GraphRouter: A Graph-based Router for LLM Selections},
  author={Feng, Tao and Shen, Yanzhen and You, Jiaxuan},
  booktitle={The Thirteenth International Conference on Learning Representations}
}

@misc{llmrouter2025,
  title        = {LLMRouter: An Open-Source Library for LLM Routing},
  author       = {Tao Feng and Haozhen Zhang and Zijie Lei and Haodong Yue and Chongshan Lin and Jiaxuan You},
  year         = {2025},
  howpublished = {\url{https://github.com/ulab-uiuc/LLMRouter}},
  note         = {GitHub repository}
}

@inproceedings{mdagent,
  title={Mdagents: An adaptive collaboration of llms for medical decision-making},
  author={Kim, Yubin and Park, Chanwoo and Jeong, Hyewon and Chan, Yik Siu and Xu, Xuhai and McDuff, Daniel and Lee, Hyeonhoon and Ghassemi, Marzyeh and Breazeal, Cynthia and Park, Hae Won},
  booktitle={The Thirty-eighth Annual Conference on Neural Information Processing Systems},
  year={2024}
}

@inproceedings{mmedagent,
  title={MMedAgent: Learning to Use Medical Tools with Multi-modal Agent},
  author={Li, Binxu and Yan, Tiankai and Pan, Yuanting and Luo, Jie and Ji, Ruiyang and Ding, Jiayuan and Xu, Zhe and Liu, Shilong and Dong, Haoyu and Lin, Zihao and others},
  booktitle={Findings of the Association for Computational Linguistics: EMNLP 2024},
  pages={8745--8760},
  year={2024}
}

@article{medrax,
  title={MedRAX: Medical Reasoning Agent for Chest X-ray},
  author={Fallahpour, Adibvafa and Ma, Jun and Munim, Alif and Lyu, Hongwei and Wang, Bo},
  journal={arXiv preprint arXiv:2502.02673},
  year={2025}
}

@article{kg4diagnosis,
  title={KG4Diagnosis: A Hierarchical Multi-Agent LLM Framework with Knowledge Graph Enhancement for Medical Diagnosis},
  author={Zuo, Kaiwen and Jiang, Yirui and Mo, Fan and Lio, Pietro},
  journal={arXiv preprint arXiv:2412.16833},
  year={2024}
}

@inproceedings{medagents,
  title={MedAgents: Large Language Models as Collaborators for Zero-shot Medical Reasoning},
  author={Tang, Xiangru and Zou, Anni and Zhang, Zhuosheng and Li, Ziming and Zhao, Yilun and Zhang, Xingyao and Cohan, Arman and Gerstein, Mark},
  booktitle={Findings of the Association for Computational Linguistics ACL 2024},
  pages={599--621},
  year={2024}
}

@misc{fallahpour2025medraxmedicalreasoningagent,
      title={MedRAX: Medical Reasoning Agent for Chest X-ray}, 
      author={Adibvafa Fallahpour and Jun Ma and Alif Munim and Hongwei Lyu and Bo Wang},
      year={2025},
      eprint={2502.02673},
      archivePrefix={arXiv},
      primaryClass={cs.LG},
      url={https://arxiv.org/abs/2502.02673}, 
}

@misc{sharma2024cxragentvisionlanguagemodelschest,
      title={CXR-Agent: Vision-language models for chest X-ray interpretation with uncertainty aware radiology reporting}, 
      author={Naman Sharma},
      year={2024},
      eprint={2407.08811},
      archivePrefix={arXiv},
      primaryClass={eess.IV},
      url={https://arxiv.org/abs/2407.08811}, 
}

@misc{chen2024visionlanguagefoundationmodelenhance,
      title={A Vision-Language Foundation Model to Enhance Efficiency of Chest X-ray Interpretation}, 
      author={Zhihong Chen and Maya Varma and Justin Xu and Magdalini Paschali and Dave Van Veen and Andrew Johnston and Alaa Youssef and Louis Blankemeier and Christian Bluethgen and Stephan Altmayer and Jeya Maria Jose Valanarasu and Mohamed Siddig Eltayeb Muneer and Eduardo Pontes Reis and Joseph Paul Cohen and Cameron Olsen and Tanishq Mathew Abraham and Emily B. Tsai and Christopher F. Beaulieu and Jenia Jitsev and Sergios Gatidis and Jean-Benoit Delbrouck and Akshay S. Chaudhari and Curtis P. Langlotz},
      year={2024},
      eprint={2401.12208},
      archivePrefix={arXiv},
      primaryClass={cs.CV},
      url={https://arxiv.org/abs/2401.12208}, 
}

@article{ong2024routellm,
  title={{RouteLLM}: Learning to route {LLMs} with preference data},
  author={Ong, Isaac and Almahairi, Amjad and Wu, Vincent and Chiang, Wei-Lin and Wu, Tianhao and Gonzalez, Joseph E and Kadous, M Waleed and Stoica, Ion},
  journal={arXiv preprint arXiv:2406.18665},
  year={2024}
}

@misc{martian,
  title = {Martian {LLM} Router},
  howpublished = {\url{https://withmartian.com/}},
}

@misc{unify,
  title = {Unify {LLM} Router},
  howpublished = {\url{https://unify.ai/}},
}

@misc{notdiamond,
  title = {NotDiamond {LLM} Router},
  howpublished = {\url{https://www.notdiamond.ai/}},
}

@inproceedings{jiang2023llm,
  title={{LLM-Blender}: Ensembling Large Language Models with Pairwise Ranking and Generative Fusion},
  author={Jiang, Dongfu and Ren, Xiang and Lin, Bill Yuchen},
  booktitle={Proceedings of the 61st Annual Meeting of the Association for Computational Linguistics (Volume 1: Long Papers)},
  year={2023}
}

@article{chen2023frugalgpt,
  title={{FrugalGPT}: How to use large language models while reducing cost and improving performance},
  author={Chen, Lingjiao and Zaharia, Matei and Zou, James},
  journal={arXiv preprint arXiv:2305.05176},
  year={2023}
}

@article{aggarwal2023automix,
  title={Automix: Automatically mixing language models},
  author={Aggarwal, Pranjal and Madaan, Aman and Anand, Ankit and Potharaju, Srividya Pranavi and Mishra, Swaroop and Zhou, Pei and Gupta, Aditya and Rajagopal, Dheeraj and Kappaganthu, Karthik and Yang, Yiming and others},
  journal={arXiv preprint arXiv:2310.12963},
  year={2023}
}

@article{stripelis2024polyrouter,
  title={TensorOpera Router: A Multi-Model Router for Efficient {LLM} Inference},
  author={Stripelis, Dimitris and Hu, Zijian and Zhang, Jipeng and Xu, Zhaozhuo and Shah, Alay and Jin, Han and Yao, Yuhang and Avestimehr, Salman and He, Chaoyang},
  journal={arXiv preprint arXiv:2408.12320},
  year={2024}
}

@article{feng2024graphrouter,
  title={GraphRouter: A Graph-based Router for {LLM} Selections},
  author={Feng, Tao and Shen, Yanzhen and You, Jiaxuan},
  journal={arXiv preprint arXiv:2410.03834},
  year={2024}
}

@article{shnitzer2023large,
  title={Large language model routing with benchmark datasets},
  author={Shnitzer, Tal and Ou, Anthony and Silva, M{\'\i}rian and Soule, Kate and Sun, Yuekai and Solomon, Justin and Thompson, Neil and Yurochkin, Mikhail},
  journal={arXiv preprint arXiv:2309.15789},
  year={2023}
}

@article{narayanan2023tryage,
  title={Tryage: Real-time, intelligent Routing of User Prompts to Large Language Models},
  author={Narayanan Hari, Surya and Thomson, Matt},
  journal={arXiv e-prints},
  year={2023}
}

@inproceedings{vsakota2024fly,
  title={Fly-swat or cannon? cost-effective language model choice via meta-modeling},
  author={{\v{S}}akota, Marija and Peyrard, Maxime and West, Robert},
  booktitle={Proceedings of the 17th ACM International Conference on Web Search and Data Mining},
  year={2024}
}

@article{srivatsa2024harnessing,
  title={Harnessing the Power of Multiple Minds: Lessons Learned from {LLM} Routing},
  author={Srivatsa, KV and Maurya, Kaushal Kumar and Kochmar, Ekaterina},
  journal={arXiv preprint arXiv:2405.00467},
  year={2024}
}

@inproceedings{yue2024largelanguagemodelcascades,
  title={Large Language Model Cascades with Mixture of Thought Representations for Cost-Efficient Reasoning},
  author={Yue, Murong and Zhao, Jie and Zhang, Min and Du, Liang and Yao, Ziyu},
  booktitle={International Conference on Learning Representations ({ICLR})},
year={2024}
}

@inproceedings{lee2024orchestrallmefficientorchestrationlanguage,
  title={{OrchestraLLM}: Efficient Orchestration of Language Models for Dialogue State Tracking},
  author={Lee, Chia-Hsuan and Cheng, Hao and Ostendorf, Mari},
  booktitle={Proceedings of the 2024 Conference of the North American Chapter of the Association for Computational Linguistics: Human Language Technologies (Volume 1: Long Papers)},
  year={2024}
}

@inproceedings{du2022glam,
  title={Glam: Efficient scaling of language models with mixture-of-experts},
  author={Du, Nan and Huang, Yanping and Dai, Andrew M and Tong, Simon and Lepikhin, Dmitry and Xu, Yuanzhong and Krikun, Maxim and Zhou, Yanqi and Yu, Adams Wei and Firat, Orhan and others},
  booktitle={International Conference on Machine Learning ({ICML})},
  year={2022},
}

@article{fedus2022switch,
  title={Switch transformers: Scaling to trillion parameter models with simple and efficient sparsity},
  author={Fedus, William and Zoph, Barret and Shazeer, Noam},
  journal={Journal of Machine Learning Research ({JMLR})},
  year={2022}
}

@article{riquelme2021scaling,
  title={Scaling vision with sparse mixture of experts},
  author={Riquelme, Carlos and Puigcerver, Joan and Mustafa, Basil and Neumann, Maxim and Jenatton, Rodolphe and Susano Pinto, Andr{\'e} and Keysers, Daniel and Houlsby, Neil},
  journal={Advances in Neural Information Processing Systems ({NeurIPS})},
  year={2021}
}

@inproceedings{shazeer2017outrageously,
  title={Outrageously Large Neural Networks: The Sparsely-Gated Mixture-of-Experts Layer},
  author={Shazeer, Noam and Mirhoseini, Azalia and Maziarz, Krzysztof and Davis, Andy and Le, Quoc and Hinton, Geoffrey and Dean, Jeff},
  booktitle={International Conference on Learning Representations},
  year={2016}
}

@article{parisi2022talm,
  title={Talm: Tool augmented language models},
  author={Parisi, Aaron and Zhao, Yao and Fiedel, Noah},
  journal={arXiv preprint arXiv:2205.12255},
  year={2022}
}

@article{tang2023toolalpaca,
  title={Toolalpaca: Generalized tool learning for language models with 3000 simulated cases},
  author={Tang, Qiaoyu and Deng, Ziliang and Lin, Hongyu and Han, Xianpei and Liang, Qiao and Cao, Boxi and Sun, Le},
  journal={arXiv preprint arXiv:2306.05301},
  year={2023}
}

@article{qin2023toolllm,
  title={Toolllm: Facilitating large language models to master 16000+ real-world apis},
  author={Qin, Yujia and Liang, Shihao and Ye, Yining and Zhu, Kunlun and Yan, Lan and Lu, Yaxi and Lin, Yankai and Cong, Xin and Tang, Xiangru and Qian, Bill and others},
  journal={arXiv preprint arXiv:2307.16789},
  year={2023}
}

@inproceedings{qintoolllm,
  title={ToolLLM: Facilitating Large Language Models to Master 16000+ Real-world APIs},
  author={Qin, Yujia and Liang, Shihao and Ye, Yining and Zhu, Kunlun and Yan, Lan and Lu, Yaxi and Lin, Yankai and Cong, Xin and Tang, Xiangru and Qian, Bill and others},
  booktitle={The Twelfth International Conference on Learning Representations}
}

@inproceedings{yaoreact,
  title={ReAct: Synergizing Reasoning and Acting in Language Models},
  author={Yao, Shunyu and Zhao, Jeffrey and Yu, Dian and Du, Nan and Shafran, Izhak and Narasimhan, Karthik R and Cao, Yuan},
  booktitle={The Eleventh International Conference on Learning Representations}
}

@article{schick2023toolformer,
  title={Toolformer: Language models can teach themselves to use tools},
  author={Schick, Timo and Dwivedi-Yu, Jane and Dess{\`\i}, Roberto and Raileanu, Roberta and Lomeli, Maria and Hambro, Eric and Zettlemoyer, Luke and Cancedda, Nicola and Scialom, Thomas},
  journal={Advances in Neural Information Processing Systems},
  volume={36},
  pages={68539--68551},
  year={2023}
}

@article{lu2024chameleon,
  title={Chameleon: Plug-and-play compositional reasoning with large language models},
  author={Lu, Pan and Peng, Baolin and Cheng, Hao and Galley, Michel and Chang, Kai-Wei and Wu, Ying Nian and Zhu, Song-Chun and Gao, Jianfeng},
  journal={Advances in Neural Information Processing Systems},
  volume={36},
  year={2024}
}

@inproceedings{
patil2024gorilla,
title={Gorilla: Large Language Model Connected with Massive {API}s},
author={Shishir G Patil and Tianjun Zhang and Xin Wang and Joseph E. Gonzalez},
booktitle={The Thirty-eighth Annual Conference on Neural Information Processing Systems},
year={2024},
url={https://openreview.net/forum?id=tBRNC6YemY}
}

@inproceedings{
chen2024advancing,
title={Advancing Tool-Augmented Large Language Models: Integrating Insights from Errors in Inference Trees},
author={Sijia Chen and Yibo Wang and Yi-Feng Wu and Qing-Guo Chen and Zhao Xu and Weihua Luo and Kaifu Zhang and Lijun Zhang},
booktitle={The Thirty-eighth Annual Conference on Neural Information Processing Systems},
year={2024},
url={https://openreview.net/forum?id=ZIpdu0cHYu}
}

@inproceedings{
rafailov2023direct,
title={Direct Preference Optimization: Your Language Model is Secretly a Reward Model},
author={Rafael Rafailov and Archit Sharma and Eric Mitchell and Christopher D Manning and Stefano Ermon and Chelsea Finn},
booktitle={Thirty-seventh Conference on Neural Information Processing Systems},
year={2023},
url={https://openreview.net/forum?id=HPuSIXJaa9}
}

@inproceedings{
liu2024apigen,
title={{APIG}en: Automated {PI}peline for Generating Verifiable and Diverse Function-Calling Datasets},
author={Zuxin Liu and Thai Quoc Hoang and Jianguo Zhang and Ming Zhu and Tian Lan and Shirley Kokane and Juntao Tan and Weiran Yao and Zhiwei Liu and Yihao Feng and Rithesh R N and Liangwei Yang and Silvio Savarese and Juan Carlos Niebles and Huan Wang and Shelby Heinecke and Caiming Xiong},
booktitle={The Thirty-eight Conference on Neural Information Processing Systems Datasets and Benchmarks Track},
year={2024},
url={https://openreview.net/forum?id=Jfg3vw2bjx}
}

@inproceedings{zhuangtoolchain,
  title={ToolChain*: Efficient Action Space Navigation in Large Language Models with A* Search},
  author={Zhuang, Yuchen and Chen, Xiang and Yu, Tong and Mitra, Saayan and Bursztyn, Victor and Rossi, Ryan A and Sarkhel, Somdeb and Zhang, Chao},
  booktitle={The Twelfth International Conference on Learning Representations}
}

@inproceedings{wu2024autogen,
  title={AutoGen: Enabling Next-Gen LLM Applications via Multi-Agent Conversation},
  author={Wu, Qingyun and Bansal, Gagan and Zhang, Jieyu and Wu, Yiran and Li, Jiale and others},
  booktitle={ICLR 2024 Workshop on Large Language Model (LLM) Agents}
}

@inproceedings{yuan2024easytool,
  title={EASYTOOL: Enhancing LLM-based Agents with Concise Tool Instruction},
  author={Yuan, Siyu and Song, Kaitao and Chen, Jiangjie and Tan, Xu and Shen, Yongliang and Ren, Kan and Li, Dongsheng and Yang, Deqing},
  booktitle={ICLR 2024 Workshop on Large Language Model (LLM) Agents}
}

@article{
DBLP:journals/tmlr/ChenM0C23,
title={Program of Thoughts Prompting: Disentangling Computation from Reasoning for Numerical Reasoning Tasks},
author={Wenhu Chen and Xueguang Ma and Xinyi Wang and William W. Cohen},
journal={Transactions on Machine Learning Research},
issn={2835-8856},
year={2023},
note={}
}

@inproceedings{gao2023pal,
  title={Pal: Program-aided language models},
  author={Gao, Luyu and Madaan, Aman and Zhou, Shuyan and Alon, Uri and Liu, Pengfei and Yang, Yiming and Callan, Jamie and Neubig, Graham},
  booktitle={International Conference on Machine Learning},
  pages={10764--10799},
  year={2023},
  organization={PMLR}
}

@inproceedings{
DBLP:conf/iclr/GouSGSYHDC24,
title={To{RA}: A Tool-Integrated Reasoning Agent for Mathematical Problem Solving},
author={Zhibin Gou and Zhihong Shao and Yeyun Gong and yelong shen and Yujiu Yang and Minlie Huang and Nan Duan and Weizhu Chen},
booktitle={The Twelfth International Conference on Learning Representations},
year={2024},
}

@inproceedings{
yuemammoth,
title={{MA}mmo{TH}: Building Math Generalist Models through Hybrid Instruction Tuning},
author={Xiang Yue and Xingwei Qu and Ge Zhang and Yao Fu and Wenhao Huang and Huan Sun and Yu Su and Wenhu Chen},
booktitle={The Twelfth International Conference on Learning Representations},
year={2024},
}

@inproceedings{chen2024towards,
  title={Towards Tool Use Alignment of Large Language Models},
  author={Chen, Zhi-Yuan and Shen, Shiqi and Shen, Guangyao and Zhi, Gong and Chen, Xu and Lin, Yankai},
  booktitle={Proceedings of the 2024 Conference on Empirical Methods in Natural Language Processing},
  pages={1382--1400},
  year={2024}
}

@inproceedings{wu2024toolplanner,
  title={ToolPlanner: A Tool Augmented LLM for Multi Granularity Instructions with Path Planning and Feedback},
  author={Wu, Qinzhuo and Liu, Wei and Luan, Jian and Wang, Bin},
  booktitle={Proceedings of the 2024 Conference on Empirical Methods in Natural Language Processing},
  pages={18315--18339},
  year={2024}
}

@article{rafailov2024direct,
  title={Direct preference optimization: Your language model is secretly a reward model},
  author={Rafailov, Rafael and Sharma, Archit and Mitchell, Eric and Manning, Christopher D and Ermon, Stefano and Finn, Chelsea},
  journal={Advances in Neural Information Processing Systems},
  volume={36},
  year={2024}
}

@article{zhao2023survey,
  title={A survey of large language models},
  author={Zhao, Wayne Xin and Zhou, Kun and Li, Junyi and Tang, Tianyi and Wang, Xiaolei and Hou, Yupeng and Min, Yingqian and Zhang, Beichen and Zhang, Junjie and Dong, Zican and others},
  journal={arXiv preprint arXiv:2303.18223},
  year={2023}
}

@inproceedings{ren2024survey,
  title={A survey of large language models for graphs},
  author={Ren, Xubin and Tang, Jiabin and Yin, Dawei and Chawla, Nitesh and Huang, Chao},
  booktitle={Proceedings of the 30th ACM SIGKDD Conference on Knowledge Discovery and Data Mining},
  pages={6616--6626},
  year={2024}
}

@inproceedings{shi2024learning,
  title={Learning to Use Tools via Cooperative and Interactive Agents},
  author={Shi, Zhengliang and Gao, Shen and Chen, Xiuyi and Feng, Yue and Yan, Lingyong and Shi, Haibo and Yin, Dawei and Ren, Pengjie and Verberne, Suzan and Ren, Zhaochun},
  booktitle={Findings of the Association for Computational Linguistics: EMNLP 2024},
  pages={10642--10657},
  year={2024}
}

@inproceedings{
liu2025toolace,
title={Tool{ACE}: Winning the Points of {LLM} Function Calling},
author={Weiwen Liu and Xu Huang and Xingshan Zeng and xinlong hao and Shuai Yu and Dexun Li and others},
booktitle={The Thirteenth International Conference on Learning Representations},
year={2025},
url={https://openreview.net/forum?id=8EB8k6DdCU}
}
\bibliographystyle{icml2026}

\newpage
\appendix
\onecolumn

\section{Related Work}
\textbf{Healthcare Agents.}
Current medical agentic systems largely follow two complementary directions: (i) collaborative reasoning setups that aggregate multiple LLM/VLM outputs via debate or voting to improve reliability \cite{mdagent,medagents,kg4diagnosis}, and (ii) tool-using agents in which an orchestrator coordinates task-specialized medical models \cite{mmedagent,medrax}. Alongside these developments, medical VLMs are increasingly being packaged into agentic pipelines for chest X-ray understanding, extending beyond perception to structured reasoning, diagnostic prediction, and report generation. For example, MedRAX \citep{fallahpour2025medraxmedicalreasoningagent} integrates multiple perception modules with multimodal LLM reasoning to answer complex queries spanning VQA and reporting, while CXR-Agent \citep{sharma2024cxragentvisionlanguagemodelschest} produces structured reports together with diagnostic outputs, and CheXagent \citep{chen2024visionlanguagefoundationmodelenhance} aims for unified chest X-ray understanding across reasoning-oriented tasks. More broadly, MDAgents \cite{mdagent} adaptively configures multi-agent collaboration based on task complexity and reports strong benchmark performance, whereas MMedAgent \cite{mmedagent} selects and composes imaging tools (e.g., segmentation, classification, report generation) across five modalities. However, as medical agents become more prevalent and increasingly tool-dependent, most work emphasizes tool \emph{orchestration} at a workflow level, while the problem of \emph{query-conditioned selection} among multiple competing tool candidates for a given subtask remains largely unexplored. This motivates our focus on tool candidate selection in this paper.

\textbf{Tool Augmented LLMs.} Tool augmented LLM has emerged as a central mechanism for extending LLM problem-solving capabilities by enabling interaction with external tools, thereby alleviating limitations such as stale knowledge and lack of real-time data access~\cite{zhao2023survey,ren2024survey}. Early work on tool-augmented language models (TALM)~\citep{parisi2022talm} demonstrated that self-play can improve reasoning and mathematical performance, while Toolformer~\citep{schick2023toolformer} showed that LLMs can self-supervise the acquisition of API-usage skills by integrating tool calls directly into text generation. As interest in tool use grew, subsequent efforts increasingly focused on constructing benchmarks and datasets to train and evaluate tool-using models, including agent-driven data generation~\citep{tang2023toolalpaca}, bootstrapping from curated seed examples~\citep{patil2024gorilla}, adapting existing datasets, and scaling supervision with strong LLMs such as GPT-4~\citep{qin2023toolllm}.

Broadly, prior work on tool learning follows two complementary directions. The first relies on prompt-based orchestration with powerful closed-source LLMs~\cite{shi2024learning,wu2024autogen,yuan2024easytool,DBLP:journals/tmlr/ChenM0C23,gao2023pal}, evolving from ReAct’s~\cite{yaoreact} thought–action prompting toward more structured multi-tool reasoning frameworks. Representative examples include Chameleon~\cite{lu2024chameleon}, which improves compositional reasoning by coordinating multiple tools with GPT-4, and ToolChain~\cite{zhuangtoolchain}, which enhances efficiency by searching the action space using A*. The second direction focuses on improving open-source LLMs through instruction tuning for tool use~\cite{DBLP:conf/iclr/GouSGSYHDC24,yuemammoth,chen2024towards,wu2024toolplanner,shi2024learning}. ToolBench~\cite{qintoolllm} marked a major step in this line by curating a large-scale dataset covering 16,464 real-world APIs and introducing ToolLLM with depth-first decision-tree trajectory annotation, albeit with later work identifying limitations due to dataset quality. Building on this foundation, TP-LLaMA~\cite{chen2024advancing} incorporates preference learning to leverage failed tool-use trajectories via DPO~\cite{rafailov2023direct,rafailov2024direct}. More recent automated data-generation frameworks, such as APIGen~\cite{liu2024apigen} and Tool-Ace~\cite{liu2025toolace}, further strengthen tool-use capabilities through multi-stage API validation and diversified query generation.

Despite this rapid progress, much of the existing literature focuses on learning to invoke individual APIs or execute short tool-use sequences. In contrast, we study tool use in medical agentic systems where tools correspond to task-specific clinical solvers (e.g., diagnosis, grounding, report generation, etc) with heterogeneous and partially overlapping capabilities; while prior work largely emphasizes learning to invoke tools or execute tool chains, it does not address query-conditioned selection among multiple competing specialist tool candidates in clinically realistic settings, which is the central focus of this paper.

\textbf{LLM Routing.} LLM routing has emerged as a central mechanism for controlling accuracy–cost trade-offs in systems that serve multiple large language models. Broadly, LLM routers act as control planes that decide which underlying model(s) should handle a given query, typically without modifying the query or the model outputs themselves. A prominent class of such systems performs prescriptive routing, where a lightweight learned classifier estimates query complexity or expected utility and routes the query to one of two (or more) candidate LLMs accordingly \cite{dinghybrid,ong2024routellm,stripelis2024polyrouter,vsakota2024fly,lee2024orchestrallmefficientorchestrationlanguage}. In this setting, routing thresholds can be calibrated on representative workloads to achieve a desired balance between response quality and inference cost.
An alternative paradigm is non-prescriptive routing, where routing decisions depend on the responses generated by candidate models rather than solely on the input query \cite{chen2023frugalgpt,aggarwal2023automix,yue2024largelanguagemodelcascades}. For example, FrugalGPT \cite{chen2023frugalgpt} employs a cascade of models ordered by cost, sequentially invoking models until a response satisfies a learned sufficiency criterion. Such approaches trade increased latency for improved cost efficiency by avoiding unnecessary invocation of stronger models.

Beyond cost-centric motivations, several routing methods focus primarily on improving response quality by exploiting complementary strengths of different models \cite{shnitzer2023large,narayanan2023tryage,feng2024graphrouter,srivatsa2024harnessing}. These methods often frame routing as a learned matching problem between queries and model capabilities, using representations of queries, models, or their interactions to guide selection. Related but distinct are ensemble-style control mechanisms, such as mixture-of-experts (MoE) architectures \cite{du2022glam,fedus2022switch,riquelme2021scaling,shazeer2017outrageously}, which dynamically select subsets of experts at the token level and combine their outputs. LLM synthesis approaches \cite{jiang2023llm} operate at the query level, routing entire inputs to multiple models and aggregating their responses. While these architectures can reduce inference costs and improve robustness, they differ fundamentally from routing systems in that they actively combine model outputs rather than selecting a single model per query. 
A primary application of LLM routing is cost reduction in LLM-powered services. Several commercial platforms, including Unify \cite{unify}, Martian \cite{martian}, and NotDiamond \cite{notdiamond}, offer routing-as-a-service solutions that transparently select an appropriate model for each query. By inserting a routing layer between applications and LLM providers, these systems report substantial cost savings - up to orders of magnitude in some deployments, while maintaining acceptable performance. Together, these developments underscore the growing importance of routing and orchestration mechanisms as foundational infrastructure for scalable LLM-based systems. Despite extensive work on LLM routing for cost and quality optimization, existing approaches largely focus on selecting among homogeneous language models and do not address the problem of query-conditioned selection among heterogeneous, task-specialized tools with partial and overlapping support. This gap motivates our focus on tool selection rather than LLM routing in this paper.

\end{document}